\documentclass[10pt,twocolumn,letterpaper]{article}
\usepackage{calrsfs}
\usepackage{cvpr}
\usepackage{times}
\usepackage{epsfig}
\usepackage{graphicx}
\usepackage{amsmath}
\usepackage{amssymb}
\usepackage{stfloats}
\usepackage{amsfonts}
\usepackage{multirow}
\usepackage{mathrsfs}
\usepackage{mathtools}
\usepackage{relsize}
\usepackage{dsfont}
\usepackage{bm}
\usepackage[caption=false]{subfig}
\usepackage{algorithm}
\usepackage{algpseudocode}
\usepackage{tablefootnote}
\DeclareMathAlphabet{\pazocal}{OMS}{zplm}{m}{n}
\newcommand{\Lb}{\pazocal{L}}
\newcommand{\Jb}{\pazocal{J}}

\usepackage[pagebackref=true,breaklinks=true,letterpaper=true,colorlinks,bookmarks=false]{hyperref}

\cvprfinalcopy 

\def\cvprPaperID{5625} 

\ifcvprfinal\fi

\newcommand*{\affaddr}[1]{#1} 
\newcommand*{\affmark}[1][*]{\textsuperscript{#1}}

\begin{document}

\title{Defending Against Universal Attacks Through Selective Feature Regeneration}
\author{\hspace*{\fill}Tejas Borkar\affmark[1]\hspace{26pt} \hfill Felix Heide\affmark[2,3] \hspace{26pt} \hfill Lina Karam\affmark[1,4]\hspace*{\fill}\\
\hspace*{\fill}
\affaddr{\affmark[1]\normalsize Arizona State University}\hspace{6pt}    
\hfill
\affaddr{\affmark[2]\normalsize Princeton University}\hspace{6pt} 
\hfill
\affaddr{\affmark[3]\normalsize Algolux}\hspace{6pt} 
\hfill
\affaddr{\affmark[4]\normalsize Lebanese American University}
\hspace*{\fill}\\
{\tt\small \{tsborkar,karam\}@asu.edu}
\hspace{8pt}
{\tt \small fheide@princeton.edu}
}

\maketitle

\begin{abstract}
Deep neural network (DNN) predictions have been shown to be vulnerable to carefully crafted adversarial perturbations. Specifically, image-agnostic (universal adversarial) perturbations added to any image can fool a target network into making erroneous predictions. Departing from existing defense strategies that work mostly in the image domain, we present a novel defense which operates in the DNN feature domain and effectively defends against such universal perturbations. Our approach identifies pre-trained convolutional features that are most vulnerable to adversarial noise and deploys trainable feature regeneration units which transform these DNN filter activations into resilient features that are robust to universal perturbations. Regenerating only the top 50\% adversarially susceptible activations in at most 6 DNN layers and leaving all remaining DNN activations unchanged, we outperform existing defense strategies across different network architectures by more than 10\% in restored accuracy. We show that without any additional modification, our defense trained on ImageNet with one type of universal attack examples effectively defends against other types of unseen universal attacks.  Code/models available at \url{https://github.com/tsborkar/Selective-feature-regeneration}
\end{abstract}

\section{Introduction}
\label{sec:intro}
 Despite the continued success and widespread use of DNNs in computer vision tasks \cite{AlexNet, vggnet, GoogLeNet, he2016deep, fastrcnn,yolo, fcn, dilated_conv}, these networks make erroneous predictions when a small magnitude, carefully crafted perturbation (adversarial noise) almost visually imperceptible to humans is added to an input image \cite{szegedy2013intriguing, adverserial_samples,madry, C&W,adv_for_GAN, deepfool, papernot, iter_FGSM, JSMA}. Furthermore, such perturbations have been successfully placed in a real-world scene via physical adversarial objects~\cite{athalye2017synthesizing, eykholt2017robust, iter_FGSM}, thus posing a security risk.
  
\begin{figure}[!t]
    \vspace{-3pt}
	\centering
	\subfloat{\includegraphics[width=0.49\textwidth]{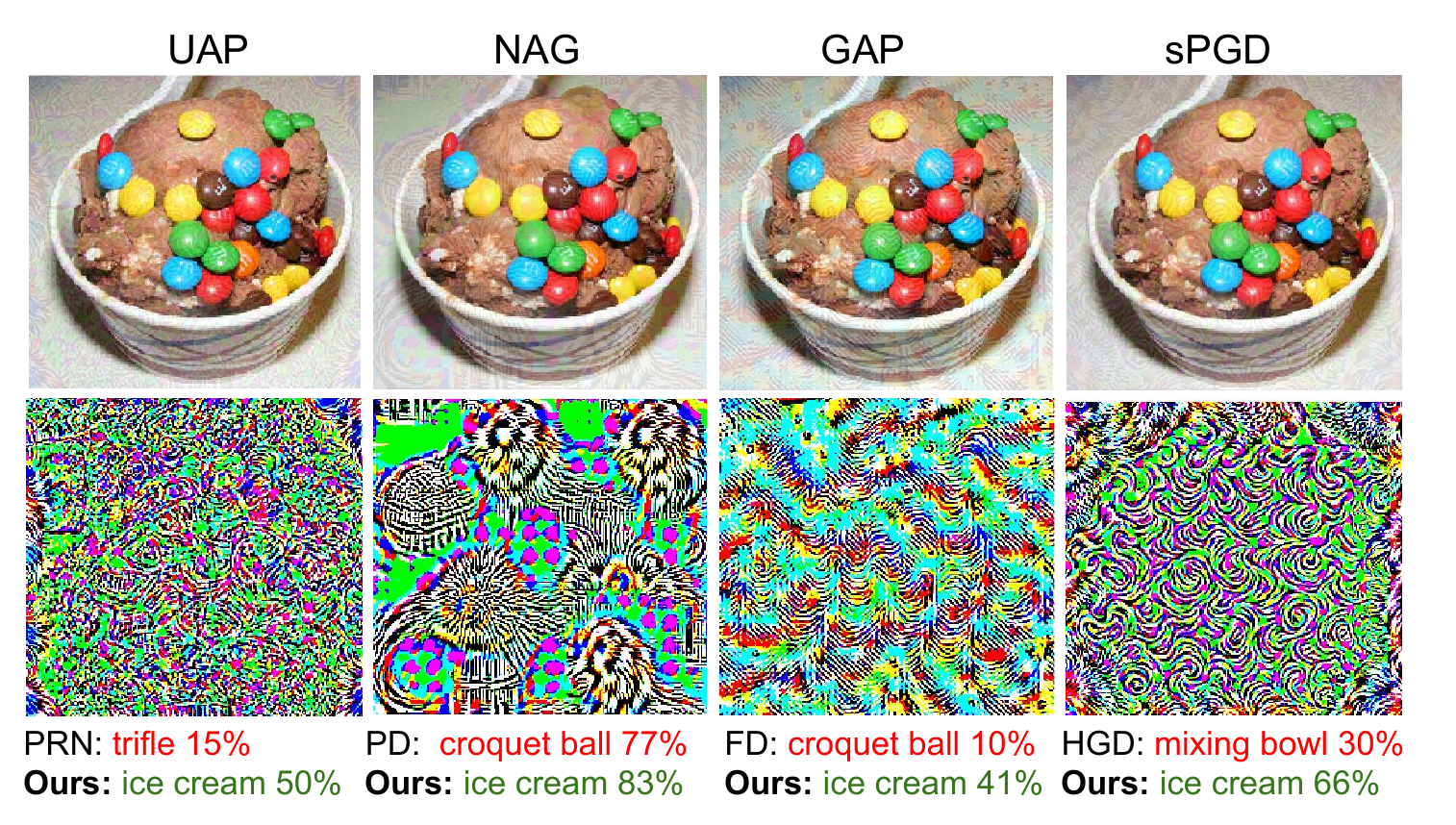}}
	\caption{Effectiveness of proposed defense against various universal perturbations: row 1 shows an image (class label: "ice cream") perturbed by different universal attacks (UAP~\cite{UAP}, NAG~\cite{nag}, GAP~\cite{gap} and sPGD~\cite{spgd}) and the second row shows the variability across different universal perturbations. Row 3 shows predictions and confidence score for the proposed defense and the next best defense (PRN~\cite{UA_def}, PD~\cite{pixel_deflection}, FD~\cite{feat_denoise} and HGD~\cite{guided_denoise}). Our method effectively defends against each universal attack by correctly classifying (green) the image with high confidence while all the other defenses misclassify it (red).}
		\label{fig:intro_fig}
\end{figure}

Most existing adversarial attacks use target network gradients to construct an image-dependent adversarial example \cite{szegedy2013intriguing, adverserial_samples, iter_FGSM, deepfool, JSMA, C&W} that has limited transferability to other networks or images \cite{szegedy2013intriguing, transfer_1, transfer_2}. Other methods to generate image-dependent adversarial samples include accessing only the network predictions~\cite{ilyas2018blackbox,black_box_1, single_pixel}, using surrogate networks \cite{papernot} and gradient approximation~\cite{obfuscated-gradients}. Although there is significant prior work on adversarial defenses such as adversarial training~\cite{szegedy2013intriguing,adverserial_samples,madry,feat_denoise}, ensemble training~\cite{ensemble_tr}, randomized image transformations and denoising~\cite{quilting, pixel_deflection, JPEG_eval, UAP_denoise,pixel_deflection, saak,JPEG_eval,Feature_dist,guided_denoise} and adversarial sample rejection~\cite{Li_2017_ICCV, safety_net, feature_squeeze, MagNet, metzen}, a DNN is still vulnerable to adversarial perturbations added to a non-negligible portion of the input~\cite{obfuscated-gradients,uesato2018adversarial}. These defenses mostly focus on making a DNN robust to image-dependent adversarial perturbations which are less likely to be encountered in realistic vision applications~\cite{UA_def,spgd}.

Our proposed work focuses on defending against universal adversarial attacks.
Unlike the aforementioned image-dependent adversarial attacks, universal adversarial attacks \cite{UAP,nag,fastfeatfool,gap,singular_UAP,spgd, Mopuri_2018_ECCV,gduap,rhp} construct a single \emph{image-agnostic} perturbation that when added to any unseen image fools DNNs into making erroneous predictions with very high confidence. These universal perturbations are also not unique and many adversarial directions may exist in a DNN's feature space (~\figurename~\ref{fig:intro_fig}, row 2)~\cite{UAP_analysis, classification_reg,robustness_dnn_geometry}.
Furthermore, universal perturbations generated for one DNN can transfer to other DNNs, making them \emph{doubly universal} \cite{UAP}. Such \emph{image-agnostic} perturbations pose a strong realistic threat model~\cite{spgd} for many vision applications as perturbations can easily be pre-computed and then inserted in real-time (in the form of a printed adversarial patch or sticker) into any scene~\cite{camera_adv,adversarial_patch}.  For example, while performing semantic segmentation, such \emph{image-agnostic} perturbations can completely hide a target class (i.e., pedestrian) in the resulting segmented scene output and adversely affect the braking action of an autonomous car \cite{UAP_seg}.

This work proposes a novel defense against a {universal adversarial} threat model~\cite{UAP,fastfeatfool,nag,gap,singular_UAP,spgd} through the following contributions: 
\begin{itemize} \vspace*{-3pt}
    \item We show the existence of a set of vulnerable convolutional filters, that are largely responsible for erroneous predictions made by a DNN in an adversarial setting and the $\ell_1$-norm of the convolutional filter weights can be used to identify such filters.  \vspace*{-3pt}
    \item Unlike, existing image-domain defenses, our proposed DNN feature space-based defense uses trainable \emph{feature regeneration units}, which regenerate activations of the aforementioned vulnerable convolutional filters into resilient features (adversarial noise masking).  \vspace*{-3pt}
    \item A fast method is proposed to generate strong synthetic adversarial perturbations for training.  \vspace*{-3pt}
    \item We extensively evaluate the proposed defense on a variety of DNN architectures and show that our proposed defense outperforms all other existing defenses~\cite{UA_def,pixel_deflection,feat_denoise, guided_denoise, madry, spgd} (~\figurename~\ref{fig:intro_fig}).   \vspace*{-3pt}
    \item Without any additional attack-specific training, our defense trained on one type of universal attack~\cite{UAP} effectively defends against other different unseen universal attacks~\cite{nag,fastfeatfool,gap,spgd,singular_UAP,gduap} (~\figurename~\ref{fig:intro_fig}) and we are the first to show such broad generalization across different universal attacks. 
\end{itemize}

\section{Related Work}
\label{sec:related_work}
Adversarial training (Adv.~tr.)~\cite{szegedy2013intriguing, adverserial_samples,madry}  has been shown to improve DNN robustness to image-dependent adversarial attacks through augmentation, in the training stage, with adversarial attack examples, which are computed on-the-fly for each mini-batch using gradient-ascent to maximize the DNN's loss. The robustness of adversarial training to black-box attacks can be improved by using perturbations computed against different target DNNs that are chosen from an ensemble of DNNs~\cite{ensemble_tr}. Kannan~{\it{et~al.}}~\cite{logit_pairing} scale adversarial training to ImageNet~\cite{imagenet} by encouraging the adversarial loss to match logits for pairs of adversarial and perturbation-free images (logit pairing) but this latter method fails against stronger iterative attacks~\cite{engstrom2018evaluating}. In addition to adversarially training the baseline DNN, prior works~(\cite{feat_denoise}, ~\cite{lamb2018fortified}) further improved DNN robustness to image-dependent attacks by denoising intermediate DNN feature maps, either through a non-local mean denoiser~(feature denoising~\cite{feat_denoise}) or a denoising auto-encoder~(fortified nets~\cite{lamb2018fortified}). Although Xie~{\it{et al.}} report effective robustness against a strong PGD attack~\cite{madry} evaluated on ImageNet~\cite{imagenet}, the additional non-local mean denoisers only add a 4\% improvement over a DNN trained using standard adversarial training. 
Compared to feature denoising~(FD)~\cite{feat_denoise}, the proposed feature regeneration approach has the following differences: (1) our \emph{feature regeneration units} are not restricted to only perform denoising, but consists of stacks of trainable convolutional layers that provide our defense the flexibility to learn an appropriate feature-restoration transform that effectively defends against universal attacks, unlike the non-local mean denoiser used in FD; (2) in a selected DNN layer, only a subset of feature maps which are the most susceptible to adversarial noise (identified by our ranking metric) are regenerated leaving all other feature maps unchanged, whereas FD denoises all feature maps, which can result in over-correction or introduce unwanted artifacts in feature maps that admit very low magnitude noise;  (3) instead of adversarially training all the parameters of the baseline DNN as in FD, we only train the parameters in the \emph{feature renegeration units} (up to 90\% less parameters than a baseline DNN) and leave all parameters in the baseline DNN unchanged, which can speed up training and reduce the risk of over-fitting.

Image-domain defenses mitigate the impact of adversarial perturbations by
utilizing non-differentiable transformations of the input such as image compression~\cite{JPEG_eval,SHEILD,Feature_dist}, frequency domain denoising~\cite{saak} and image quilting and reconstruction~\cite{quilting, UAP_denoise} etc. However, such approaches introduce unnecessary artifacts in clean images resulting in accuracy loss \cite{UA_def}\cite{pixel_deflection}. Prakash {\it{et~al.}} \cite{pixel_deflection} propose a two-step defense that first performs random local pixel redistribution, followed by a wavelet denoising. Liao~{\it{et al.}}~\cite{guided_denoise} append a denoising autoencoder at the input of the baseline DNN and train it using a reconstruction loss that minimizes the error between higher layer representations of the DNN for an input pair of clean and denoised adversarial images (high level guided denoiser). Another popular line of defenses explores the idea of first detecting an adversarially perturbed input and then either abstaining from making a prediction or further pre-processing adversarial input for reliable predictions \cite{Li_2017_ICCV, safety_net, feature_squeeze, MagNet, metzen}.

All of the aforementioned defenses are geared towards image-specific gradient-based attacks and none of them has, as of yet, been shown to defend against \emph{image-agnostic} attacks. Initial attempts at improving robustness to universal attacks involved modelling the distribution of such perturbations~\cite{UAP,generative_uap, adversarial_game}, followed by model fine-tuning over this distribution of universal perturbations. However, the robustness offered by these methods has been unsatisfactory~\cite{spgd,UAP} as the retrained network ends up overfitting to the small set of perturbations used. Extending adversarial training for image-dependent attacks to universal attacks has been attempted in \cite{spgd} and \cite{shafahi2018universal}. Ruan and Dai~\cite{uap_detect} use additional shadow classifiers to identify and reject images perturbed by universal perturbations. Akhtar {\it{et~al.}} \cite{UA_def} propose a defense against the universal adversarial perturbations attack~(UAP)~\cite{UAP}, using a detector which identifies adversarial images and then denoises them using a learnable \emph{Perturbation Rectifying Network} (PRN).

\section{Universal threat model}
\label{sec:problem}
Let $\mu_c$ represent the distribution of clean (unperturbed) images in $\mathbb{R}^d$, $\mathcal{F}(\cdot)$ be a classifier that predicts a class label $\mathcal{F}(x)$ for an image $x \in \mathbb{R}^d$. The universal adversarial perturbation attack seeks a perturbation vector $v \in \mathbb{R}^d$ under the following constraints \cite{UAP}:
\begin{equation}
\vspace*{-1ex}
    \underset{x\sim\mu_c}{P}\Big(\mathcal{F}(x+v) \neq \mathcal{F}(x)\Big) \geq (1-\delta) \;\: \textrm{s.t.} \;\: \|v\|_p \leq \xi
\end{equation}
where $P(\cdot)$ denotes probability, $\|\cdot\|_p$ is the $\ell_p$-norm with $p \in [1,\infty)$, $(1-\delta)$ is the target \emph{fooling ratio} with $\delta \in [0,1)$ (i.e., the fraction of samples in $\mu_c$ that change labels when perturbed by an adversary), and $\xi$ controls the magnitude of adversarial perturbations. 

\section{Feature-Domain Adversarial Defense}
\label{sec:proposed_approach}
\subsection{Stability of Convolutional Filters}
\label{subsec:filter_search}

\begin{figure}[!t]
  \vspace*{-2ex}
	\centering
	\hspace*{-3pt}\subfloat[]{ \includegraphics[width=0.48\textwidth, height=1in]{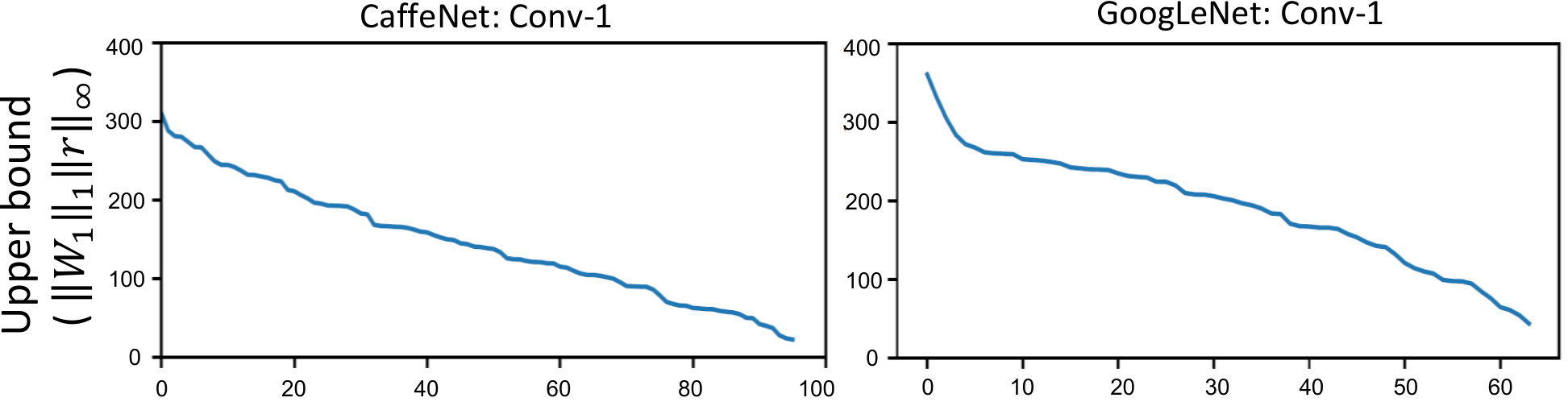}
	\vspace*{-3ex}
	\label{subfig:upper_bound}}
	\vspace*{-10pt}
	\hspace*{-3pt}\subfloat[]{ \includegraphics[width=0.48\textwidth, height=1in]{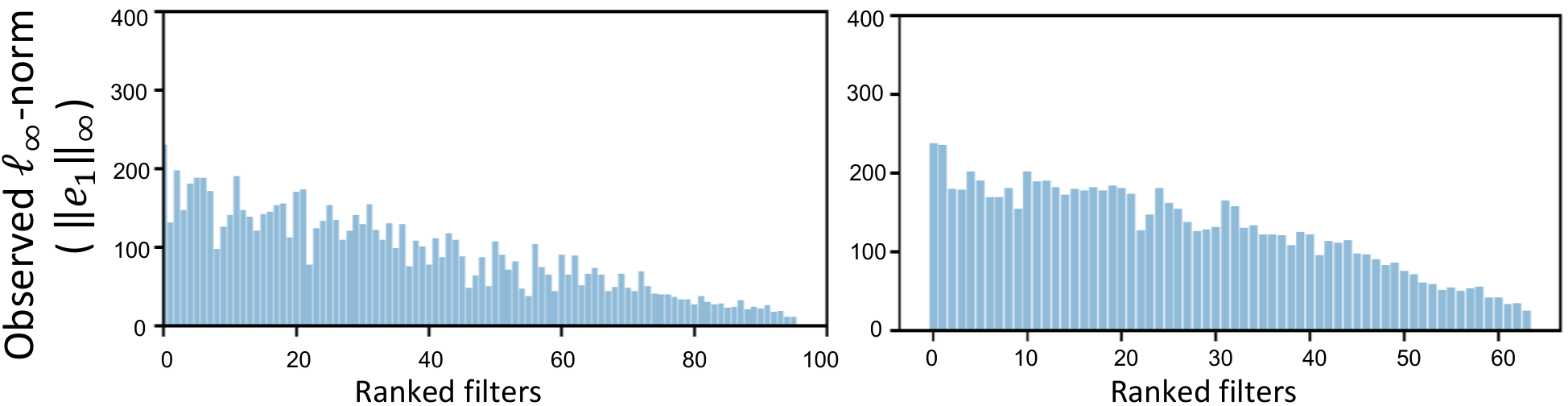}
	\label{subfig:observed_noise}}   
	\caption{Observed $\ell_\infty$-norm for {universal adversarial} noise in the activation maps of ranked convolutional filters (ordered using our $\ell_1$-norm ranking measure, from most to least vulnerable) of the first layer of CaffeNet \cite{AlexNet} and GoogLeNet \cite{GoogLeNet}. The $\ell_\infty$-norm attack is used with $\xi \leq 10$, i.e. $\|{\bf{r}}\|_\infty \leq 10$. (a) Adversarial noise upper-bound (Equation~\ref{eqn:simpler}) in ranked conv-1 filter activations of DNNs. (b) Observed $\ell_\infty$-norm for adversarial noise in ranked conv-1 filter activations of DNNs. }  
	\label{fig:rank_f1}
	\vspace*{-1ex}
\end{figure}

\begin{figure}[!t]
	\centering
	\hspace*{-6ex}
    \includegraphics[width=0.42\textwidth,height=1.5in]{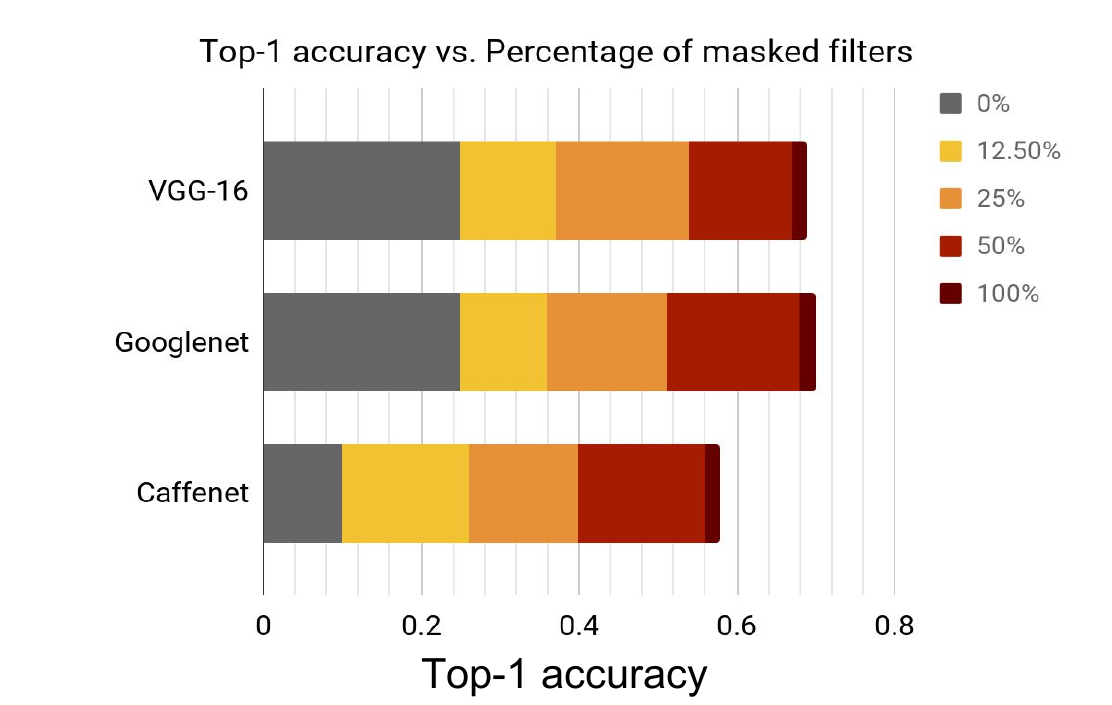}
	\caption{Effect of masking $\ell_\infty$-norm universal adversarial noise in ranked convolutional filter activations of the first layer in CaffeNet \cite{AlexNet}, GoogLeNet \cite{GoogLeNet} and VGG-16 \cite{vggnet}, evaluated on a 1000-image subset of the ImageNet~\cite{imagenet} training set. Top-1 accuracies for perturbation-free images are 0.58, 0.70 and 0.69 for CaffeNet, GoogLeNet and VGG-16, respectively. Similarly, top-1 accuracies for adversarially perturbed images with no noise masking are 0.1, 0.25 and 0.25 for CaffeNet, GoogLeNet and VGG-16, respectively. Masking the noise in just 50\% of the ranked filter activations restores most of the lost accuracy for all three DNNs.}   
	\vspace*{-1ex}
	\label{fig:masking}
\end{figure}

In this work, we assess the vulnerability of individual convolutional filters and show that, for each layer, certain filter activations are significantly more disrupted than others, especially in the early layers of a DNN.

For a given layer, let $\phi_m(u)$ be the output (activation map) of the $m^{\textrm{th}}$ convolutional filter with kernel weights $W_m$ for an input $u$. Let $e_m = \phi_m(u + r) - \phi_m(u)$ be the additive noise (perturbation) that is caused in the output activation map $\phi_m(u)$  as a result of applying an additive perturbation $r$ to the input $u$. It can be shown (refer to Supplementary Material) that $e_m$ is bounded as follows: 
\begin{equation}
\vspace*{-1ex}
     \|e_m\|_\infty \leq \|W_m\|_1\|r\|_p
     \label{eqn:simpler}
\end{equation}
where as before $\|\cdot\|_p$ is the $\ell_p$-norm with $p \in [1, \infty)$.
Equation~\ref{eqn:simpler} shows that the $\ell_1$-norm of the filter weights can be used to identify and rank convolutional filter activations in terms of their ability to restrict perturbation in their activation maps. For example, filters with a small weight $\ell_1$-norm would result in insignificant small perturbations in their output when their input is perturbed, and are thus considered to be less vulnerable to perturbations in the input. For an $\ell_\infty$-norm universal adversarial input, Figure~\ref{subfig:upper_bound} shows the upper-bound on the adversarial noise in ranked (using the proposed $\ell_1$-norm ranking) conv-1 filter activations of CaffeNet \cite{AlexNet} and GoogLeNet \cite{GoogLeNet}, while \figurename~\ref{subfig:observed_noise} shows the corresponding observed $\ell_\infty$-norm for adversarial noise in the respective DNN filter activations. We can see that our $\|W\|_1$-based ranking correlates well with the degree of perturbation (maximum magnitude of the noise perturbation) that is induced in the filter outputs. Similar observations can be made for other convolutional layers in the network. 

In ~\figurename~\ref{fig:masking}, we evaluate the impact of masking the adversarial noise in such ranked filters on the overall top-1 accuracy of CaffeNet \cite{AlexNet}, VGG-16 \cite{vggnet} and GoogLeNet \cite{GoogLeNet}. Specifically, we randomly choose a subset of 1000 images (1 image per class) from the ImageNet \cite{imagenet} training set and generate adversarially perturbed images by adding an $\ell_\infty$-norm universal adversarial perturbation~\cite{UAP}. The top-1 accuracies for perturbation-free images are 0.58, 0.70 and 0.69 for CaffeNet, GoogLeNet and VGG-16, respectively. Similarly, the top-1 accuracies for adversarially perturbed images of the same subset are 0.10, 0.25 and 0.25 for CaffeNet, GoogLeNet and VGG-16, respectively. Masking the adversarial perturbations in 50\% of the most vulnerable filter activations significantly improves DNN performance, resulting in top-1 accuracies of 0.56, 0.68 and 0.67 for CaffeNet, GoogLeNet and VGG-16, respectively, and validates our proposed selective feature regeneration scheme. See Figure 1 in Supplementary Material for similar experiments for higher layers.

\subsection{Resilient Feature Regeneration Defense}
\label{subsec:defender_unit}
\begin{figure*}[!t]
	\centering
	\vspace*{-1ex}
	\includegraphics[width=0.9\textwidth]{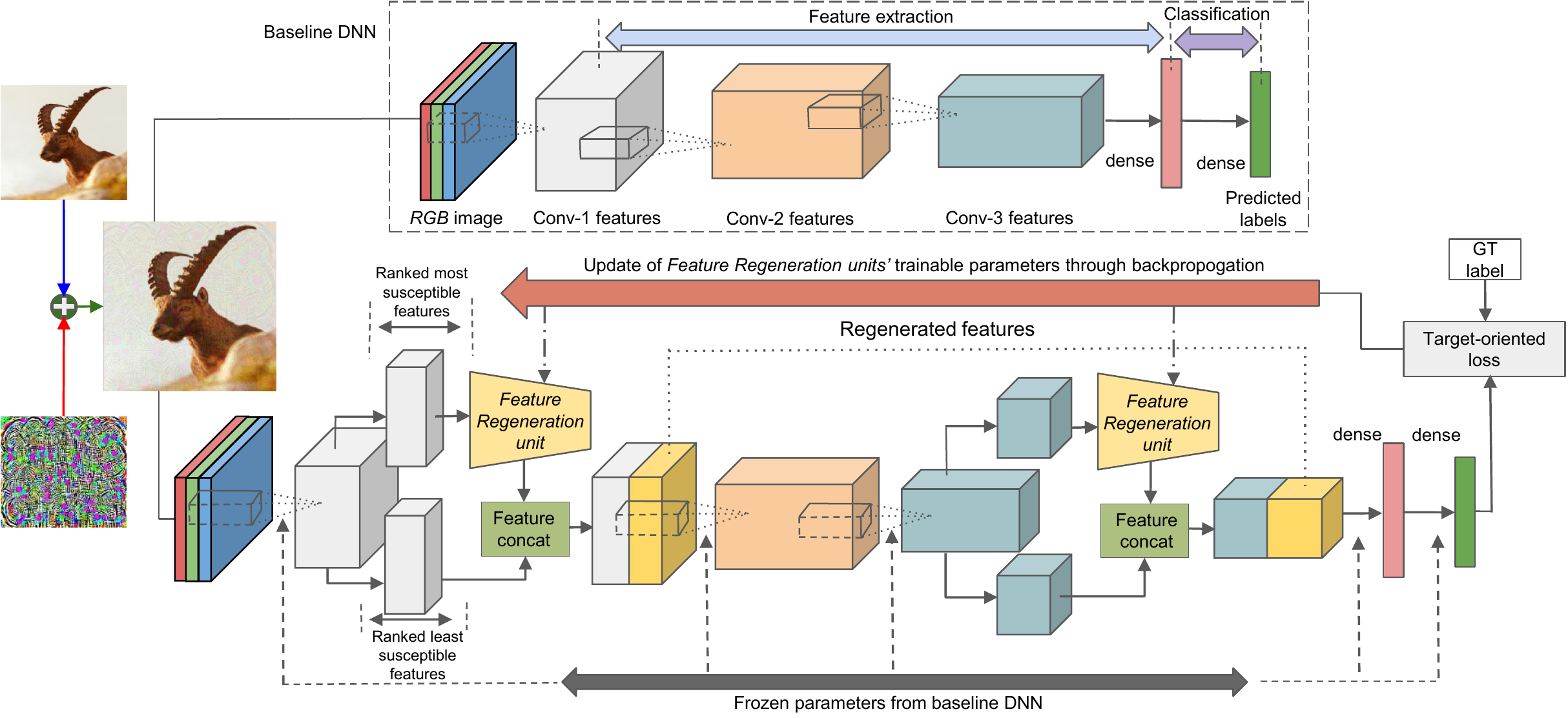}
	\caption{{\bf{Resilient Feature Regeneration Defense}}: Convolutional filter activations in the baseline DNN (top) are first sorted in order of vulnerability to adversarial noise using their respective filter weight norms (Section~\ref{subsec:filter_search}). For each considered layer, we use  a \emph{feature regeneration unit}, consisting of a residual block with a single skip connection (4 layers), to regenerate only the most adversarially susceptible activations into resilient features that restore the lost accuracy of the baseline DNN, while leaving the remaining filter activations unchanged. We train these units on both clean and perturbed images in every mini-batch using the same target loss as the baseline DNN such that all parameters of the baseline DNN are left unchanged during training.}   
	\vspace*{-1ex}
	\label{fig:blockD_fig}
\end{figure*}
Our proposed defense is illustrated in Figure~\ref{fig:blockD_fig}. We learn a task-driven feature restoration transform (i.e., \emph{feature regeneration unit}) for convolutional filter activations severely disrupted by adversarial input. Our \emph{feature regeneration unit} does not modify the remaining activations of the baseline DNN. A similar approach of learning corrective transforms for making networks more resilient to image blur and additive white Gaussian noise has been explored in~\cite{deepcorrect}. 

\begin{figure*}[!t]
	\centering
	\vspace*{-2ex}
	\includegraphics[width=0.99\textwidth, height=2in]{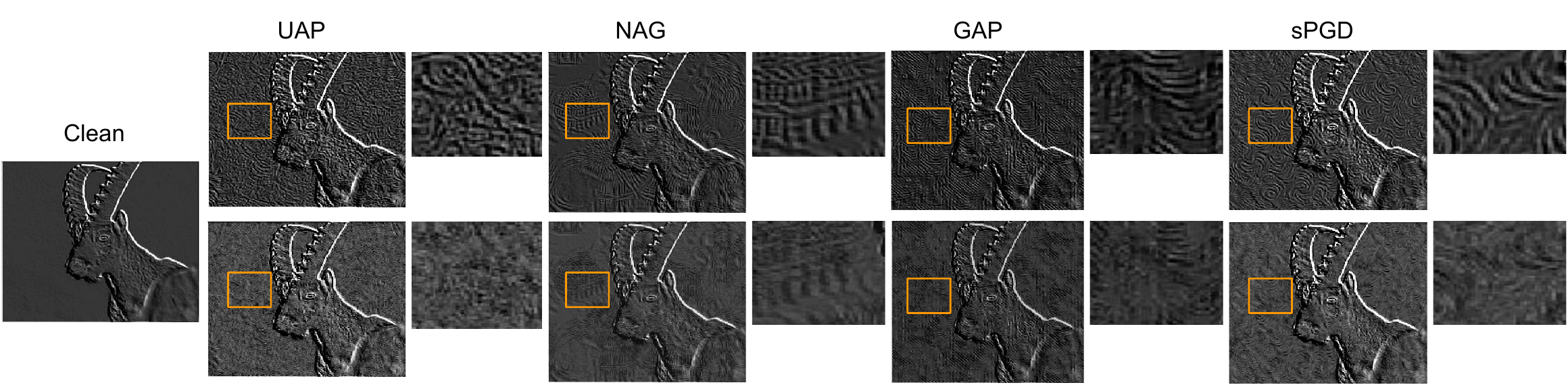}
	\caption{Effectiveness of \emph{feature regeneration units} at masking adversarial perturbations in DNN feature maps for images perturbed by universal perturbations~(UAP~\cite{UAP}, NAG~\cite{nag}, GAP~\cite{gap} and sPGD~\cite{spgd}). Perturbation-free feature map (clean), different adversarially perturbed feature maps (Row 1) and corresponding feature maps regenerated by \emph{feature regeneration units} (Row 2) are obtained for a single filter channel in conv1\_1 layer of VGG-16~\cite{vggnet}, along with an enlarged view of a small region in the feature map (yellow box). \emph{Feature regeneration units} are only trained on UAP~\cite{UAP} attack examples but are very effective at suppressing adversarial artifacts generated by unseen attacks (e.g., NAG~\cite{nag}, GAP~\cite{gap} and sPGD~\cite{spgd}).
	}   
	\vspace*{-2ex}
	\label{fig:feat_corr}
\end{figure*}

Let $S_l$ represent a set consisting of indices for convolutional filters in the $l^{th}$ layer of a DNN. Furthermore, let $S_{l_{reg}}$ be the set of indices for filters we wish to regenerate (Section~\ref{subsec:filter_search}) and let $S_{l_{adv}}$ be the set of indices for filters whose activations are not regenerated (i.e., $S_l = S_{l_{reg}} \cup S_{l_{adv}}$). If $\Phi_{S_{l_{reg}}}$ represents the convolutional filter outputs to be regenerated in the $l^{th}$ layer, then our \emph{feature regeneration unit} in layer $l$ performs a feature regeneration transform $\mathcal{D}_{l}(\cdot)$ under the following conditions:
\begin{equation}
\mathcal{D}_l(\Phi_{S_{l_{reg}}}(u+r)) \approx \Phi_{S_{l_{reg}}}(u)
\label{eqn:adv_corr}
\end{equation}
and 
\begin{equation}
\mathcal{D}_l(\Phi_{S_{l_{reg}}}(u)) \approx \Phi_{S_{l_{reg}}}(u)
\label{eqn:ori_corr}
\end{equation}
where $u$ is the unperturbed input to the $l^{th}$ layer of convolutional filters and $r$ is an additive perturbation that acts on $u$. In Equations~\ref{eqn:adv_corr} and~\ref{eqn:ori_corr}, $\approx$ denotes similarity based on classification accuracy in the sense that features are restored to regain the classification accuracy of the original perturbation-free activation map. Equation~\ref{eqn:adv_corr} forces $\mathcal{D}_{l}(\cdot)$ to pursue task-driven feature regeneration that restores lost accuracy of the DNN while Equation~\ref{eqn:ori_corr} ensures that prediction accuracy on unperturbed activations is not decreased, without any additional adversarial perturbation detector. We implement $\mathcal{D}_l(\cdot)$ (i.e., \emph{feature regeneration unit}) as a shallow \emph{residual block}~\cite{he2016deep}, consisting of two stacked $3\times3$ convolutional layers sandwiched between a couple of $1\times1$ convolutional layers and a single skip connection. $\mathcal{D}_l(\cdot)$ is estimated using a target loss from the baseline network, through backpropagation, see Figure~\ref{fig:blockD_fig}, but with significantly fewer trainable parameters compared to the baseline network. 

Given an $L$ layered DNN $\Phi$, pre-trained for an image classification task, $\Phi$ can be represented as a function that maps network input $x$ to an $N$-dimensional output label vector $\Phi(x)$ as follows:
\begin{equation}
\Phi = \Phi_{L}\circ\Phi_{L-1}\circ\ldots\Phi_{2}\circ\Phi_{1}
\end{equation}
where $\Phi_{l}$ is a mapping function (set of convolutional filters, typically followed by a non-linearity) representing the $l^{th}$ DNN layer and $N$ is the dimensionality of the DNN's output (i.e., number of classes). Without any loss of generality, the resulting DNN after deploying a \emph{feature regeneration unit} that operates on the set of filters represented by $S_{l_{reg}}$ in layer $l$ is given by:
\begin{equation}
\Phi_{reg} = \Phi_{L}\circ\Phi_{L-1}\circ\ldots\Phi_{l_{reg}} \ldots\Phi_{2}\circ\Phi_{1}
\label{eqn:new_dnn}
\end{equation}
where $\Phi_{l_{reg}}$ represents the new mapping function for layer $l$, such that $\mathcal{D}_{l}(\cdot)$ regenerates only activations of the filter subset $\Phi_{S_{l_{reg}}}$ and all the remaining filter activations (i.e., $\Phi_{S_{l_{adv}}}$) are left unchanged. If $\mathcal{D}_l(\cdot)$ is parameterized by ${\bf{\theta}}_l$, then the \emph{feature regeneration unit} can be trained by minimizing:
\begin{equation}
\small
\Jb({\bf{\theta}}_l) = \frac{1}{K}\mathlarger{\sum}\limits_{k=1}^{K}\Lb(y_k,\Phi_{reg}(x_{k}))
\label{eqn:target_loss}
\end{equation}
\normalsize
where $\Lb$ is the same target loss function of the baseline DNN (e.g., cross-entropy classification loss), $y_k$ is the target output label for the $k^{th}$ input image $x_{k}$, $K$ represents the total number of images in the training set consisting of both clean and perturbed images. As we use both clean and perturbed images during training, $x_{k}$ in Equation~\ref{eqn:target_loss}, represents a clean or an adversarially perturbed image. 

In~\figurename~\ref{fig:feat_corr}, we visualize DNN feature maps perturbed by various universal perturbations and the corresponding feature maps regenerated by our \emph{feature regeneration units}, which are only trained on UAP~\cite{UAP} attack examples. Compared to the perturbation-free feature map (clean), corresponding feature maps for adversarially perturbed images (Row 1) have distinctly visible artifacts that reflect the universal perturbation pattern in major parts of the image. In comparison, feature maps regenerated by our \emph{feature regeneration units} (Row 2) effectively suppress these adversarial perturbations, preserve the object discriminative attributes of the clean feature map and are also robust to unseen attacks (e.g, NAG~\cite{nag}, GAP~\cite{gap} and sPGD~\cite{spgd}), as illustrated in~\figurename~\ref{fig:feat_corr} and Table~\ref{table:cross_attack}.

\subsection{Generating Synthetic Perturbations}
\label{subsec:synthetic_data}
Training-based approaches are susceptible to data overfitting, especially when the training data is scarce or does not have adequate diversity. Generating a diverse set of adversarial perturbations ( $\geq$ 100) using existing attack algorithms (e.g., \cite{UAP,nag,gap, spgd}), in order to avoid overfitting, can be computationally prohibitive. We propose a fast method (Algorithm~\ref{alg:gen_sample}) to construct synthetic universal adversarial perturbations from a small set of adversarial perturbations, $V \subseteq \mathbb{R}^d$, that is computed using any existing universal attack generation method (\cite{UAP, nag, gap, spgd}). Starting with the synthetic perturbation $v_{syn}$ set to zero, we iteratively select a random perturbation $v_{new} \in V$  and a random scale factor $\alpha \in [0,1]$  and update $v_{syn}$ as follows:
\begin{equation}
v_{syn}(t) = \alpha v_{new} + (1-\alpha) v_{syn}(t-1)
\label{eqn:syn_update}
\end{equation}
where $t$ is the iteration number. This process is repeated until the $\ell_2$-norm of $v_{syn}$ exceeds a threshold $\eta$. We set the threshold $\eta$ to be the minimum $\ell_2$-norm of perturbations in the set $V$. 

Unlike the approach of Akhtar {\it{et~al.}}~\cite{UA_def}, which uses an iterative random walk along pre-computed adversarial directions, the proposed algorithm has two distinct advantages: 1) the same algorithm can be used for different types of attack norms without any modification, and 2) Equation~\ref{eqn:syn_update} (Step 5 in Algorithm~\ref{alg:gen_sample}) automatically ensures that the $\ell_\infty$-norm of the perturbation does not violate the constraint for an $\ell_\infty$-norm attack (i.e., $\ell_\infty$-norm $\leq \xi$) and, therefore, no additional steps, like computing a separate perturbation unit vector and ensuring that the resultant perturbation strength is less than $\xi$, are needed.
\begin{figure}[]
\vspace*{-3ex}
\begin{algorithm}[H]
\caption{Generating Synthetic Adversarial Perturbation} 
\label{alg:gen_sample}
\begin{algorithmic}[1]
\small
\renewcommand{\algorithmicrequire}{\textbf{Input:}}
 \renewcommand{\algorithmicensure}{\textbf{Output:}}
\Require Set of pre-computed perturbations $V \subseteq \mathbb{R}^d$ such 
that~$v_i \in V$ is the $i^{th}$ perturbation; threshold $\eta$
\Ensure Synthetic perturbation $v_{syn} \in \mathbb{R}^d$
\State{$v_{syn}$ = 0}
\While {$\|{\bf{v}}_{syn}\|_2 \leq \eta$}
\State {$\alpha \thicksim \text{uniform}(0,1)$ 
\State $v_{new}\overset{\scriptsize\textrm{rand}}{\large\thicksim} V$}
\State $v_{syn} = \alpha v_{new} + (1-\alpha) v_{syn}$
\EndWhile
\State \textbf{return} $v_{syn}$
\end{algorithmic}
\end{algorithm}
\vspace*{-6ex}
\end{figure}

\section{Assessment}
\vspace*{-3pt}
\label{sec:results}
We use the ImageNet validation set (ILSVRC2012) \cite{imagenet} with all 50000 images and a single crop evaluation (unless specified otherwise) in our experiments. All our experiments are implemented using Caffe \cite{caffe} and for each tested attack we use publicly provided code. We report our results in terms of top-1 accuracy and the restoration accuracy proposed by Akhtar {\it{et~al.}}~\cite{UA_def}. Given a set $I_c$ containing clean images and a set $I_{p/c}$ containing clean and perturbed images in equal numbers, the restoration accuracy is given by:
\vspace{-8pt}
\begin{equation}
    \textrm{Restoration accuracy} =  \frac{\textrm{acc}(I_{p/c})}{\textrm{acc}(I_c)} 
    \label{eqn:res_acc}
\vspace{-4pt}
\end{equation}
where acc($\cdot$) is the top-1 accuracy.
We use the universal adversarial perturbation~(UAP)~attack~\cite{UAP} for evaluation (unless specified otherwise) and compute 5 independent {universal adversarial} test perturbations per network using a set of 10000 held out images randomly chosen from the ImageNet training set with the fooling ratio for each perturbation lower-bounded to 0.8 on the held out images and the maximum normalized inner product between any two perturbations for the same DNN upper-bounded to 0.15. 

\subsection{Defense Training Methodology}
\label{subsec:defense_train}
In our proposed defense (Figure~\ref{fig:blockD_fig}), only the parameters for \emph{feature regeneration units} have to be trained and these parameters are updated to minimize the cost function given by Equation~\ref{eqn:target_loss}.
Although we expect the prediction performance of defended models to improve  with  higher regeneration ratios (i.e., fraction of convolutional filter activations regenerated), we only regenerate 50\% of the convolutional filter activations in a layer and limit the number of deployed \emph{feature regeneration units} (1 per layer) as $\min(\# \textrm{DNN layers}, 6)$\footnote{From ~\figurename~\ref{fig:masking} (main paper) and Figure 1 in Supplementary Material, we observe that an empirical regeneration ratio of 50\% works well. Similarly, although \emph{feature regeneration units} can be deployed for each layer in a DNN, from Figure 2 in Supplementary Material, we observe that regenerating features in at most 6 layers in a DNN effectively recovers lost prediction performance.}. Using Algorithm~\ref{alg:gen_sample}, we generate 2000 synthetic perturbations from a set $V$ of 25 original perturbations \cite{UAP} and train \emph{feature regeneration units} on a single Nvidia Titan-X using a standard SGD optimizer, momentum of 0.9 and a weight decay of 0.0005 for 4 epochs of the ImageNet training set \cite{imagenet}. The learning rate is dropped by a factor of 10 after each epoch with an initial learning rate of 0.1. After a defense model has been trained as outlined above, we can further iterate through the training of our defense with additional adversarial perturbations computed against our defense, which ensures robustness to secondary attacks against our defense~(Section~\ref{subsubsec:secondary_attack}). 


\subsection{Analysis and Comparisons}
\label{subsec: comp&res}
\subsubsection{Robustness across DNN Architectures}
\label{subsubsec:cross-dnn}
Top-1 accuracy of adversarially perturbed test images for various DNNs (no defense) and our proposed defense for respective DNNs is reported in Table~\ref{table:cross-dnn} under both white-box (same network used to generate and test attack) and black-box (tested network is different from network used to generate attack) settings. As {universal adversarial perturbations} can be \emph{doubly universal}, under a black-box setting, we evaluate a target DNN defense (defense is trained for attacks on target DNN) against a perturbation generated for a different network. Top-1 accuracy for baseline DNNs is severely affected by both white-box and black-box attacks, whereas our proposed defense is not only able to effectively thwart the white-box attacks but is also able to generalize to attacks constructed for other networks without further training (Table~\ref{table:cross-dnn}). Since different DNNs can share common adversarial directions in their feature space, our \emph{feature regeneration units} learn to regularize such directions against unseen data and, consequently, to defend against black-box attacks. 

\vspace{-4pt}
\subsubsection{Robustness across Attack Norms}
\label{subsubsec:attack_norm}

\begin{table}[] 
\caption{{\textbf{Cross-DNN evaluation on ILSVRC2012}}: Top-1 accuracy against a $\ell_\infty$-norm {UAP~\cite{UAP} attack} with $\xi = 10$  and  target fooling ratio of 0.8. DNNs in column one are tested with attacks generated for DNNs in row one.}
\centering
\vspace{4pt}
\resizebox{0.4\textwidth}{!}{
\renewcommand{\arraystretch}{1.}%
\begin{tabular}{cccccc}
\hline
\multicolumn{1}{c}{} & {\textbf{CaffeNet}} & {\textbf{VGG-F }}  & {\textbf{GoogleNet }} & {\textbf{VGG-16 }} & {\textbf{Res152} } \\ \hline
\multicolumn{6}{c}{\multirow{2}{*}{CaffeNet \cite{AlexNet}, orginal accuracy 56.4\%}}                       \vspace*{-2pt}                                     \\
\multicolumn{6}{c}{}                                                                                                              \\ \hline
\multicolumn{1}{c|}{CaffeNet}       & 0.109            & 0.298          &0.456              & 0.431         &  0.405             \\
\multicolumn{1}{c|}{Ours}           & \textbf{0.542}    & \textbf{0.524}  & \textbf{0.510}    &  \textbf{0.457}   & \textbf{0.470}      \\ \hline
\multicolumn{6}{c}{\multirow{2}{*}{VGG-F \cite{vgg-f}, original accuracy 58.4\%}}                           \vspace*{-2pt}                                       \\
\multicolumn{6}{c}{}                                                                                                              \\ \hline
\multicolumn{1}{c|}{VGG-F}          & 0.299            & 0.150            &0.461           & 0.417            &  0.426             \\
\multicolumn{1}{c|}{Ours}           & \textbf{0.556}    & \textbf{0.550}   &\textbf{0.548}     &  \textbf{0.492} & \textbf{0.513}      \\ \hline
\multicolumn{6}{c}{\multirow{2}{*}{GoogLeNet \cite{GoogLeNet}, original accuracy 68.6\%}}                            \vspace*{-2pt}                                  \\
\multicolumn{6}{c}{}                                                                                                              \\ \hline
\multicolumn{1}{c|}{GoogLeNet}      & 0.519           & 0.539         &0.260              & 0.472            &    0.473            \\
\multicolumn{1}{c|}{Ours}           & \textbf{0.651}   & \textbf{0.653} &\textbf{0.653}     &\textbf{0.637}  & \textbf{0.642}      \\ \hline
\multicolumn{6}{c}{\multirow{2}{*}{VGG-16 \cite{vggnet}, original accuracy 68.4\%}}                                     \vspace*{-2pt}                            \\
\multicolumn{6}{c}{}                                                                                                              \\ \hline
\multicolumn{1}{c|}{VGG-16}         & 0.549            & 0.559         & 0.519                & 0.240          &   0.484            \\
\multicolumn{1}{c|}{Ours}           & \textbf{0.615}   & \textbf{0.622} &\textbf{0.646}     & \textbf{0.655} & \textbf{0.631}      \\ \hline
\multicolumn{6}{c}{\multirow{2}{*}{Res152 \cite{he2016deep}, original accuracy 79\%}}                                     \vspace*{-2pt}                             \\
\multicolumn{6}{c}{}                                                                                                              \\ \hline
\multicolumn{1}{c|}{Res152}          & 0.720                  & 0.726               &  0.692               &0.626                &  0.270             \\
\multicolumn{1}{c|}{Ours}           & \textbf{0.764}                   & \textbf{0.769}                & \textbf{0.769}     &  \textbf{0.763}               &\textbf{0.761}                \\ \hline
\end{tabular}
}
\label{table:cross-dnn} 
\end{table}

\begin{table}[]
\centering
\caption{{\bf{Same-norm evaluation on ILSVRC2012}}: Restoration accuracy of DNNs and defenses against an $\ell_\infty$-norm UAP~\cite{UAP} attack with $\xi = 10$.}
\vspace{4pt}
\resizebox{0.45\textwidth}{!}{
\renewcommand{\arraystretch}{1.1}%
\begin{tabular}{cccccc}
\hline
\textbf{Methods} & \textbf{CaffeNet} & \textbf{VGG-F} & \textbf{GoogLeNet} & \textbf{VGG-16} & \textbf{Res152} \\ \hline   
\multicolumn{6}{c}{$\ell_\infty$-norm attack, $\xi =  10$ }   \\ \hline
Baseline      & 0.596             & 0.628          & 0.691              & 0.681           & 0.670              \\
PRN \cite{UA_def}                      & 0.936             & 0.903          & 0.956              & 0.690              &  0.834             \\
PRN+det \cite{UA_def}             & 0.952             & 0.922          & 0.964          &   0.690            & 0.834              \\
PD \cite{pixel_deflection}                    & 0.873            & 0.813          & 0.884              & 0.818           & 0.845               \\

JPEG comp. \cite{JPEG_eval}  &0.554 &0.697 &0.830 &0.693 & 0.670\\

Feat. Distill. \cite{Feature_dist} &0.671 &0.689 &0.851 &0.717 &0.676\\

HGD~\cite{guided_denoise} &n/a &n/a &n/a &n/a &0.739\\

Adv. tr.~\cite{madry} & n/a &n/a &n/a &n/a &0.778 \\

FD~\cite{feat_denoise} &n/a &n/a &n/a &n/a &0.819 \\

Ours                      & \textbf{0.976}    & \textbf{0.967} & \textbf{0.970}     & \textbf{0.963}  &  \textbf{0.982}              \\ \hline
\end{tabular}
}
\vspace*{-3pt}
\label{table:comp_res}

\end{table}

\begin{table}[]
\centering
\caption{ {\textbf{Cross-norm evaluation on ILSVRC2012}}: Restoration accuracy against an $\ell_2$-norm UAP~\cite{UAP} attack. Our defense, as well as the other defense models, are trained only on $\ell_\infty$-norm attack examples with $\xi=10$. }
\vspace{4pt}
\resizebox{0.45\textwidth}{!}{
\renewcommand{\arraystretch}{1.1}%
\begin{tabular}{cccccc}
\hline
\textbf{Methods}    & \textbf{CaffeNet}    & \textbf{VGG-F}    & \textbf{GoogLeNet}    & \textbf{VGG-16}   & \textbf{Res152} \\ \hline
\multicolumn{6}{c}{$\ell_2$-norm attack, $\xi$ = 2000} \\ \hline
Baseline            & 0.677                     &  0.671                 &0.682                       &0.697     & 0.709              \\
PRN  \cite{UA_def}               & 0.922                     & 0.880                  & 0.971                      & 0.834          &  0.868           \\
PRN+det \cite{UA_def}            & 0.936                     &  0.900                 & {\bf{0.975}}                      & 0.835         &   0.868       \\
PD  \cite{pixel_deflection}             & 0.782                     &0.784                   & 0.857                      & 0.809        & 0.840          \\
HGD~\cite{guided_denoise} &n/a &n/a &n/a &n/a &0.730 \\

Adv. tr.~\cite{madry} &n/a &n/a & n/a &n/a & 0.778 \\

FD~\cite{feat_denoise} &n/a &n/a &n/a &n/a &0.818 \\
Ours                &{\bf{0.964}}                      & {\bf{0.961}} & 0.912                       & {\bf{0.876}}             & {\bf{0.926}}     \\ \hline
\end{tabular}}
\vspace*{-3pt}
\label{table:cross_norm}
\end{table}

\begin{table}[]
\caption{Restoration accuracy on ILSVRC2012 for $\ell_\infty$-norm UAP~\cite{UAP} attack with stronger perturbation strengths ($\xi$) against CaffeNet. Our defense, as well as the other defense models, are trained only on $\ell_\infty$-norm attack examples with $\xi$=10.}
\vspace{4pt}
\centering
\resizebox{0.3\textwidth}{!}{
\renewcommand{\arraystretch}{1.1}%
\begin{tabular}{ccccc}
\hline
\textbf{Method} &\textbf{$\xi=10$} &\textbf{$\xi=15$} & \textbf{$\xi=20$} & \textbf{$\xi=25$}\\ \hline
Baseline      & 0.596               & 0.543            &0.525          &0.519          \\
PRN~\cite{UA_def}  &0.936           &0.603             &0.555          &0.526 \\
PRN+det~\cite{UA_def} &0.952        &0.604             &0.555           &0.526 \\
PD~\cite{pixel_deflection}&0.873     &0.616             &0.549           &0.524 \\
Ours        &\textbf{0.976}       &\textbf{0.952}    &\textbf{0.896} &\textbf{0.854}          
 \\ \hline
\end{tabular}
}
\vspace*{-3ex}
\label{table:stronger_pert}
\end{table}

Here, we evaluate defense robustness against both $\ell_\infty$-norm and $\ell_2$-norm~UAP~\cite{UAP} attacks. Since an effective defense must not only recover the DNN accuracy against adversarial images but must also maintain a high accuracy on clean images, we use restoration accuracy (Equation~\ref{eqn:res_acc}) to measure adversarial defense robustness (Tables~\ref{table:comp_res} and \ref{table:cross_norm}). While Akhtar {\it{et~al.}}~\cite{UA_def} (PRN and PRN+det) only report defense results on the UAP~\cite{UAP} attack, we also compare results with pixel-domain defenses such as Pixel Deflection (PD~\cite{pixel_deflection}) and High Level Guided Denoiser (HGD~\cite{guided_denoise}), defenses that use JPEG compression (JPEG comp.~\cite{JPEG_eval}) or DNN-based compression like Feature Distillation~(Feat. Distill.~\cite{Feature_dist}), defenses that use some variation of adversarial training like Feature Denoising~(FD~\cite{feat_denoise}) and standard Adversarial training~(Adv. tr.~\cite{madry}).

In Table~\ref{table:comp_res}, we report results for an $\ell_\infty$-norm UAP attack~\cite{UAP} against various DNNs and show that our proposed defense outperforms all the other defenses\footnote{FD~\cite{feat_denoise}, HGD~\cite{guided_denoise} and Adv. tr.~\cite{madry} defenses publicly provide trained defense models only for Res152~\cite{he2016deep}) among the evaluated DNNs; we report results using only the DNN models provided by the respective authors.} for all networks with the highest restoration accuracy (98.2\%) being achieved for Res152~\cite{he2016deep}. Our \emph{feature regeneration units} are trained on $\ell_\infty$-norm attack examples (same-norm evaluation). Even without a perturbation detector, our defense outperforms the existing defense with a perturbation detector (PRN+det) of Akhtar {\it{et~al.}}~\cite{UA_def} for all networks. Similarly, for Res152~\cite{he2016deep}, we outperform adversarially trained defenses~(FD~\cite{feat_denoise}, Adv. tr.~\cite{madry}) and pixel denoisers (PD~\cite{pixel_deflection}, HGD~\cite{guided_denoise}) by more than 10\%. In Table~\ref{table:cross_norm}, we also evaluate how well our defense trained on an $\ell_\infty$-norm attack defends against an $\ell_2$-norm attack (cross-norm evaluation). Our \emph{feature regeneration units} are able to effectively generalize to even cross-norm attacks and outperform all other defenses for most DNNs.

\vspace{-6pt}
\subsubsection{Stronger Attack Perturbations ($\xi > 10$)}
\label{subsubsec:stronger_attacks}
Although we use an attack perturbation strength $\xi$ = 10 during training, in Table~\ref{table:stronger_pert}, we evaluate the robustness of our defense when the adversary violates the attack threat model using a higher perturbation strength. Compared to the baseline DNN (no defense) as well as PRN~\cite{UA_def} and PD~\cite{pixel_deflection}, our proposed defense is much more effective at defending against stronger perturbations, outperforming other defenses by almost 30\% even when the attack strength is more than double the value used to train our defense. Although defense robustness decreases for unseen higher perturbation strengths, our defense handles this drop-off much more gracefully and shows much better generalization across attack perturbation strengths, as compared to existing defenses. We also note that adversarial perturbations are no longer visually imperceptible at $\xi = 25$.

\vspace{-8pt}
\subsubsection{Generalization to Unseen Universal Attacks}
\label{subsubsec:other_attacks}
Although the proposed method effectively defends against UAP~\cite{UAP} attacks (Tables\ref{table:cross-dnn}-\ref{table:stronger_pert}), we also assess its robustness to other unseen universal attacks without additional attack-specific training. Note that \cite{UA_def} and \cite{spgd} do not cover this experimental setting. Since existing attacks in the literature are tailored to specific DNNs, we use CaffeNet~\cite{AlexNet} and Res152~\cite{he2016deep} DNNs for covering a variety of universal attacks like Fast Feature Fool (FFF)~\cite{fastfeatfool}, Network for adversary generation (NAG)~\cite{nag}, Singular fool (S.Fool)~\cite{singular_UAP}, Generative adversarial perturbation (GAP)~\cite{gap}, Generalizable data-free universal adversarial perturbation~(G-UAP)~\cite{gduap}, and stochastic PGD (sPGD)~\cite{spgd}. 

Our defense trained on just UAP~\cite{UAP} attack examples is able to effectively defend against all other universal attacks and outperforms all other existing defenses (Table~\ref{table:cross_attack}). Even against stronger universal attacks like NAG~\cite{nag} and GAP~\cite{gap}, we outperform all other defenses including PRN~\cite{UA_def}, which is also trained on similar UAP~\cite{UAP} attack examples, by almost 10\%. From our results in Table~\ref{table:cross_attack}, we show that our \emph{feature regeneration units} learn transformations that generalize effectively across perturbation patterns (\figurename~\ref{fig:feat_corr}). Note that we are the first to show such broad generalization across universal attacks. 

\begin{table}[]
\centering
\caption{ {\textbf{Robustness to unseen attacks}}: Restoration accuracy evaluated on ILSVRC2012, against other unseen universal attacks using our defense trained on just $\ell_\infty$-norm UAP~\cite{UAP} attack examples with a fooling ratio and $\ell_\infty$-norm of 0.8 and 10, respectively. Results for all other defenses are reported using publicly provided defense models. Attacks are constructed for the baseline DNN.}
\vspace{6pt}
\resizebox{0.48\textwidth}{!}{
\renewcommand{\arraystretch}{1.1}%
\begin{tabular}{c|ccc|ccc}
\hline
\multicolumn{1}{c}{}& \multicolumn{3}{|c}{\textbf{CaffeNet}}&\multicolumn{3}{|c}{\textbf{Res152}} \\ \hline
\textbf{Methods}    & \textbf{FFF}~\cite{fastfeatfool}    & \textbf{NAG}~\cite{nag}    & \textbf{S.Fool}~\cite{singular_UAP}    & \textbf{GAP}~\cite{gap}  &\textbf{G-UAP}~\cite{gduap} & \textbf{sPGD}~\cite{spgd} \\ \hline

Baseline &0.645 &0.670   &0.815    &0.640  & 0.726  & 0.671              \\
PRN  \cite{UA_def}  &0.729 &0.660    & 0.732 & 0.774     &0.777     & 0.823            \\
PD  \cite{pixel_deflection} & 0.847  &0.767 & 0.871 & 0.784  &0.807       &0.890           \\
HGD~\cite{guided_denoise} &n/a &n/a &n/a &0.663  &0.782  &0.932 \\

Adv. tr~\cite{madry} &n/a &n/a & n/a &0.776 &0.777   &0.775  \\

FD~\cite{feat_denoise} &n/a &n/a &n/a &0.815 &0.813  &0.815 \\
Ours &{\bf{0.941}}  & {\bf{0.840}}  &{\bf{0.914}} & {\bf{0.922}}     &{\bf{0.914}}        &{\bf{0.976}}      \\ \hline
\end{tabular}}
\vspace*{-2ex}
\label{table:cross_attack}
\end{table}

\vspace{-6pt}
\subsubsection{Robustness to Secondary White-Box Attacks}
\label{subsubsec:secondary_attack}
Although in practical situations, an attacker may not have full or even partial knowledge of a defense, for completeness, we also evaluate our proposed defense against a white-box attack on the defense (secondary attacks), i.e., adversary has full access to the gradient information of our \emph{feature regeneration units}. We use the UAP~\cite{UAP} (on CaffeNet) and sPGD~\cite{spgd} (on Res152) attacks for evaluation. 

\begin{figure}[]
	\centering
	\includegraphics[width=0.45\textwidth]{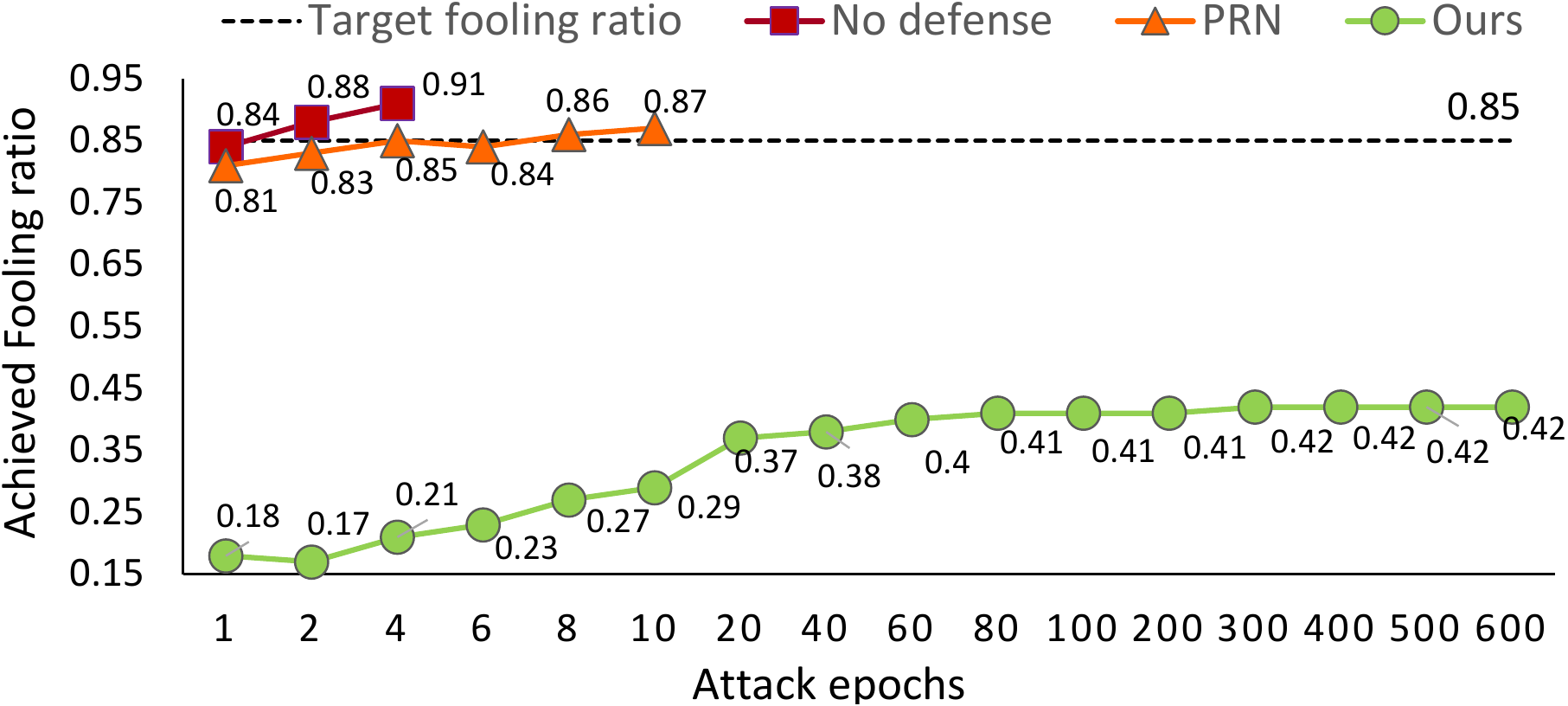}
	\caption{{\bf{Robustness to white-box attacks against defense (secondary attacks)}}: Achieved fooling ratio by attacker vs. attack epochs for an $\ell_\infty$-norm UAP~\cite{UAP} attack ($\xi = 10$) against CaffeNet~\cite{AlexNet}, where the attacker has full knowledge of the baseline DNN and also the defense. The target fooling ratio for attack is set to 0.85. }	
	\label{fig:type2_attack}
\end{figure}

Figure~\ref{fig:type2_attack} shows the robustness of our defense to such a secondary UAP~\cite{UAP} attack seeking to achieve a target fooling ratio of 0.85 on our defense for the CaffeNet~\cite{AlexNet} DNN. Such an attack can easily converge (achieve target fooling ratio) against a baseline DNN in less than 2 attack epochs, eventually achieving a final fooling ratio of 0.9. Similarly, we observe that even PRN~\cite{UA_def} is susceptible to a secondary UAP~\cite{UAP} attack, achieving a fooling ratio of 0.87, when the adversary can access gradient information for its \emph{Perturbation Rectifying Network}. In comparison, using our defense model with iterative adversarial example training (as described in Section~\ref{subsec:defense_train}), the white-box adversary can achieve a maximum fooling ratio of only 0.42, which is 48\% lower than the fooling ratio achieved against PRN~\cite{UA_def}, even after attacking our defense for 600 attack epochs.
Similarly, in Table~\ref{table:white_spgd}, using the same attack setup outlined in~\cite{spgd}, we evaluate white-box sPGD~\cite{spgd} attacks computed by utilizing gradient-information of both the defense and the baseline DNNs, for Res152~\cite{he2016deep}. As shown in Table~\ref{table:white_spgd}, our defense trained using sPGD attack examples computed against both the baseline DNN and our defense, is robust to subsequent sPGD white-box attacks. 

\begin{table}[]
\caption{Top-1 accuracy for white-box $\ell_\infty$-norm sPGD~\cite{spgd} attack against Res152-based $\ell_\infty$-norm defenses ($\xi = 10$), evaluated on ILSVRC2012. Top-1 accuracy for Res152 on clean images is 0.79. }
\vspace{4pt}
\centering
\resizebox{0.45\textwidth}{!}{
\renewcommand{\arraystretch}{1.1}%

\begin{tabular}{cccccc}
\hline
\textbf{Baseline} &\textbf{Ours} & \textbf{FD}~\cite{feat_denoise} & \textbf{Adv. tr.}~\cite{madry} & \textbf{HGD}~\cite{guided_denoise}& \textbf{Shared tr.}~\cite{spgd}\tablefootnote{ As an implementation of Shared Adversarial Training (Shared tr.~\cite{spgd}) was not publicly available, we report results published by the authors in~\cite{spgd} and which were only provided for white-box attacks computed against the defense, whereas results for white-box attacks against the baseline DNN were not provided.} \\ \hline

 0.270 &\textbf{0.731}  &0.641  &0.635  &0.689 &  0.727              \\ \hline
\end{tabular}
}
\vspace*{-2ex}
\label{table:white_spgd}
\end{table}

\section{Conclusion}
\label{sec:conclusion}
We show that masking adversarial noise in a few select DNN activations significantly improves their adversarial robustness. To this end, we propose a novel selective feature regeneration approach that effectively defends against {universal perturbations}, unlike existing adversarial defenses which either pre-process the input image to remove adversarial noise and/or retrain the entire baseline DNN through adversarial training. We show that the $\ell_1$-norm of the convolutional filter kernel weights can be effectively used to rank convolutional filters in terms of their susceptibility to adversarial perturbations. Regenerating only the top 50\% ranked adversarially susceptible features in a few DNN layers is enough to restore DNN robustness and outperform all existing defenses. We validate the proposed method by comparing against existing state-of-the-art defenses and show better generalization across different DNNs, attack norms and even unseen attack perturbation strengths. In contrast to existing approaches, our defense trained solely on one type of universal adversarial attack examples effectively defends against other unseen universal attacks, without additional attack-specific training. We hope this work encourages researchers to design adversarially robust DNN architectures and training methods which produce convolutional filter kernels that have a small $\ell_1$-norm. 


{\small
\bibliographystyle{ieee_fullname}
\bibliography{myref}
}
\clearpage
\section*{Supplementary Material}
\maketitle
 \setcounter{section}{0}
 \def\thesection{\Alph{section}}

\section{Maximum Adversarial Perturbation}
We show in Section~\ref{subsec:filter_search} of the main paper that the maximum possible adversarial perturbation in a convolutional filter activation map is proportional to the $\ell_1$-norm of its corresponding filter kernel weights. Here, we provide a proof for Equation~\ref{eqn:simpler} in the main paper. For simplicity but without loss of generality, let $A$ be a single-channel $n\times n$ input to a $k\times k$ convolutional filter with kernel $W$. For illustration, consider a 3$\times$3 input $A$ and a 2$\times$2 kernel $W$ as shown below:

\begin{center}
$A = \begin{bmatrix}
   a_1 & a_2 & a_3 \\
   a_4 & a_5 & a_6 \\
   a_7 & a_8 & a_9
\end{bmatrix}$ and $W = \begin{bmatrix}
    w_{11} & w_{12} \\
    w_{21} & w_{22} \\
\end{bmatrix}$ \\
\end{center}
Assuming the origin for the kernel $W$ is at the top-left corner and no padding for $A$ (same proof applies also if padding is applied), then the vectorized convolutional output

\begin{center} ${\bf{e}}=vec(A*W)$ \\
                        $= \begin{bmatrix} 
                        w_{11}\cdot a_1 + w_{12} \cdot a_2 + w_{21} \cdot a_4 + w_{22} \cdot a_5 \\
                        w_{11}\cdot a_2 + w_{12} \cdot a_3 + w_{21} \cdot a_5 + w_{22} \cdot a_6 \\
                        w_{11} \cdot a_4 + w_{12} \cdot a_5 + w_{21} \cdot a_7 + w_{22} \cdot a_8 \\
                        w_{11} \cdot a_5 + w_{12} \cdot a_6 + w_{21} \cdot a_8 + w_{22} \cdot a_9
                        \end{bmatrix}$  
                        \\ \vspace{3ex}
    
\end{center}
can be expressed as a matrix-vector product as follows:
\begin{equation}
                        {\bf{e}} = vec(A*W) = M{\bf{r}} 
\end{equation}
            \begin{equation}
             \label{eqn:conv_mat}
                     \footnotesize
                      M = \begin{bmatrix}
                        w_{11} \hspace{-1ex}&w_{12}\hspace{-1ex} &0\hspace{-1ex} &w_{21}\hspace{-1ex} &w_{22}\hspace{-1ex}  &0\hspace{-1ex} &0\hspace{-1ex} &0\hspace{-1ex} &0 \\
                        0\hspace{-1ex} &w_{11}\hspace{-1ex} &w_{12}\hspace{-1ex} &0\hspace{-1ex} &w_{21}\hspace{-1ex} &w_{22}\hspace{-1ex} &0\hspace{-1ex} &0\hspace{-1ex} &0 \\
                        0\hspace{-1ex} &0\hspace{-1ex} &0\hspace{-1ex} &w_{11}\hspace{-1ex} &w_{12}\hspace{-1ex} &0\hspace{-1ex} &w_{21}\hspace{-1ex} &w_{22}\hspace{-1ex} &0 \\
                        0\hspace{-1ex} &0\hspace{-1ex} &0\hspace{-1ex} &0\hspace{-1ex} &w_{11}\hspace{-1ex} &w_{12}\hspace{-1ex} &0\hspace{-1ex} &w_{21}\hspace{-1ex} &w_{22} 
                        \end{bmatrix} 
                    \end{equation}
                    
                    \begin{equation} 
                    {\bf{r}}^T=\begin{bmatrix} a_1 \hspace{-1ex}&a_2 \hspace{-1ex} &a_3 \hspace{-1ex} &a_4\hspace{-1ex} &a_5\hspace{-1ex} &a_6 \hspace{-1ex}&a_7 \hspace{-1ex}&a_8\hspace{-1ex} &a_9\end{bmatrix}
                \end{equation}
\normalsize
where $vec(\cdot)$ unrolls all elements of the input matrix with $N_1$ rows and $N_2$ columns into an output column vector of size $N_1N_2$, $M$ is a circulant convolution matrix formed using the elements of $W$ and ${\bf{r}}= vec(A)$. 

Similarly, for $A \in \mathbb{R}^{n\times n}$ and $W \in \mathbb{R}^{k\times k}$ such that $w_{ij}$ is an element in row $i$ and column $j$ of $W$, we have $M \in \mathbb{R}^{(n-k+1)^2 \times n^2}$, and $e \in \mathbb{R}^{{(n-k+1)}^2}$ is given by:
\begin{equation}
    e = M{\bf{r}} 
\end{equation}
\vspace*{-3ex}
\begin{equation}
    \|e\|_\infty = \|M{\bf{r}}\|_\infty = \max\limits_{1\leq i \leq  {(n-k+1)}^2} | \sum\limits_{j=1}^{n^2} m_{ij}r_{j}|
\end{equation}
\begin{center} $ \leq \max\limits_{1\leq i \leq (n-k+1)^2} \sum\limits_{j=1}^{n^2}|m_{ij}||r_j|
$ \\
$  \hspace{11ex} \leq \Bigg( \max\limits_{1 \leq i \leq (n-k+1)^2}\sum\limits_{j=1}^{n^2}|m_{ij}| \Bigg) \max\limits_{1 \leq j \leq n^2}|r_j|
$ \\
\end{center}
\begin{equation}
\hspace{6ex} \leq \Bigg( \max\limits_{1 \leq i \leq (n-k+1)^2}\sum\limits_{j=1}^{n^2}|m_{ij}| \Bigg)\|{\bf{r}}\|_\infty
\label{eqn:l1_norm_ori}
\end{equation} \\
where ${\bf{r}} = vec(A),  {\bf{r}} \in \mathbb{R}^{n^2}$ such that $r_j$ is the $j^\text{th}$ element in the vector ${\bf{r}}$ and $m_{ij}$ is the element in row $i$ and column $j$ of the matrix $M$. 

From Equation~\ref{eqn:conv_mat},  $\sum\limits_{j=1}^{n^2}|m_{ij}|$ is always equal to the $\ell_1$-norm of the filter kernel weights $\|W\|_1 = \sum\limits_{i'=1}^{k}\sum\limits_{j'=1}^{k}|w_{i'j'}|$ for any row $i$, $1 \leq i \leq (n-k+1)^2$. Equation~\ref{eqn:l1_norm_ori}, can now be rewritten as:
\begin{equation}
    \|{\bf{e}}\|_\infty \leq \|W\|_1\|{\bf{r}}\|_\infty
\end{equation}
Since $\|\cdot\|_\infty \leq \|\cdot\|_1$ and $\|\cdot\|_\infty \leq \|\cdot\|_2$, we have the following inequality:
\begin{equation}
    \|{\bf{e}}\|_\infty \leq \|W\|_1 \|{\bf{r}}\|_p
\end{equation}
where $p = 1, 2, \infty$.

\begin{figure*}[!t]
	\centering
    \hspace*{4ex}
    \subfloat{\includegraphics[width=0.46\textwidth, height=1.8in]{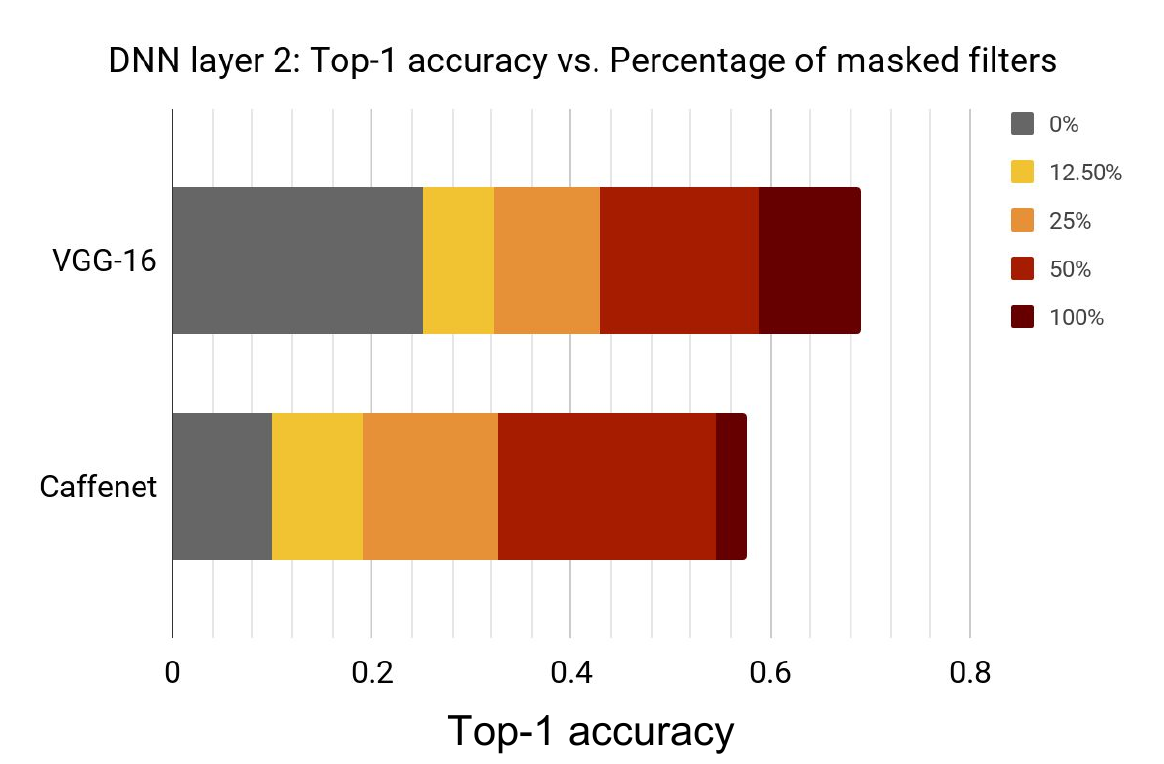}}
    \subfloat{\includegraphics[width=0.46\textwidth, height=1.8in]{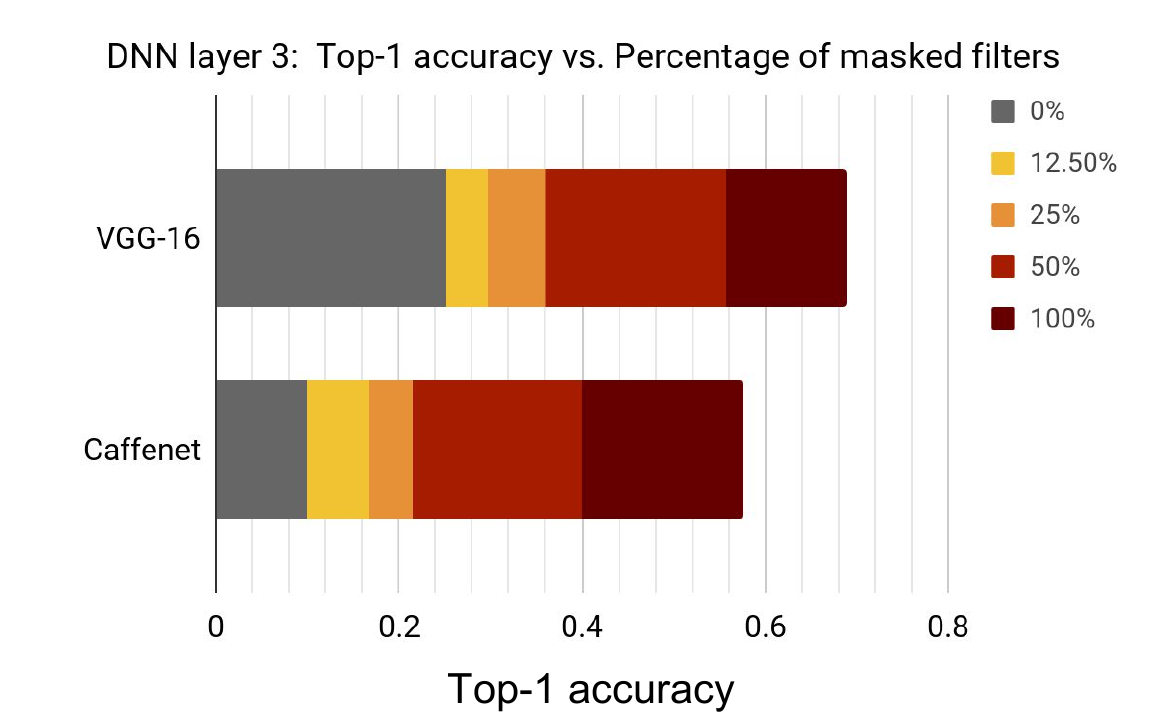}} \hfill
    \vspace*{-1ex}
    \subfloat{\includegraphics[width=0.46\textwidth, height=1.8in]{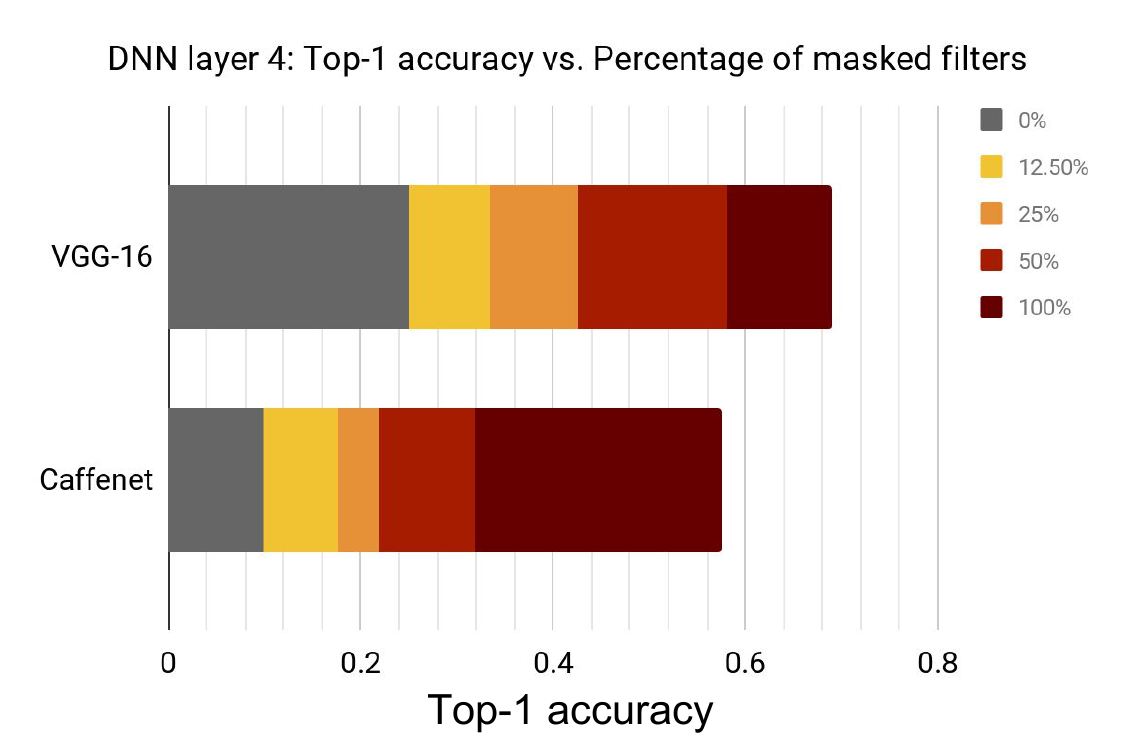}}
    \subfloat{\includegraphics[width=0.46\textwidth, height=1.8in]{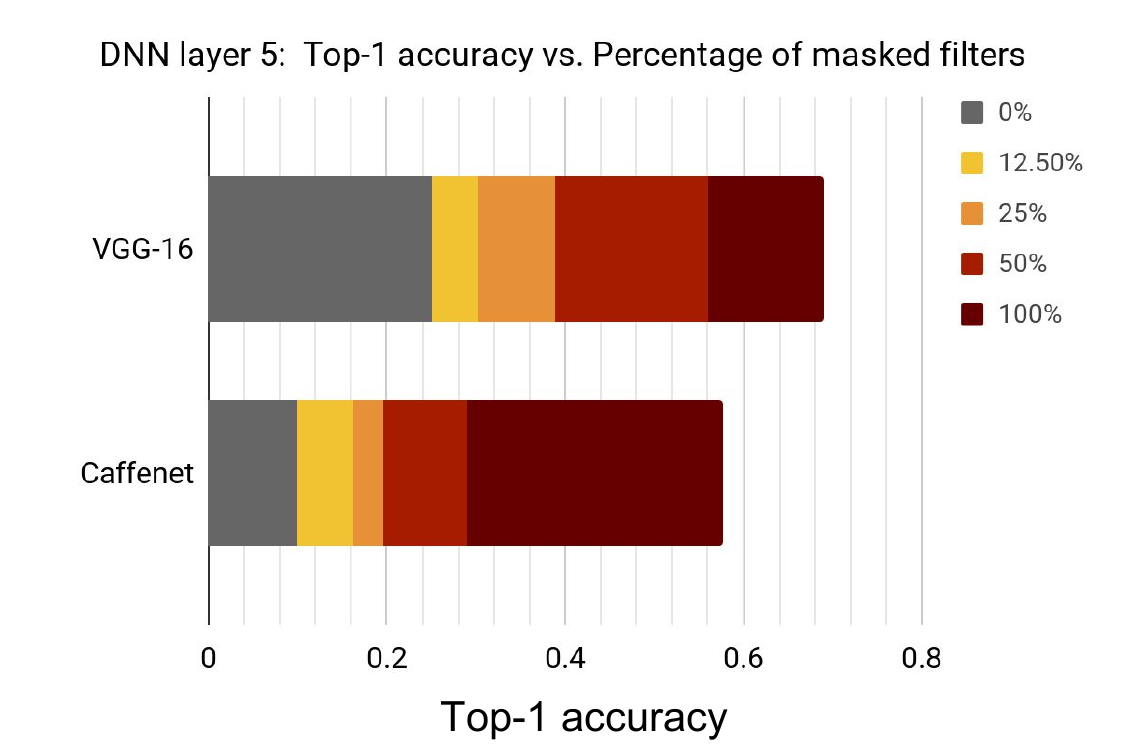}} \hfill
	\caption{Effect of masking $\ell_\infty$-norm universal adversarial noise in ranked convolutional filter activations of CaffeNet \cite{AlexNet} and VGG-16 \cite{vggnet}, evaluated on a 1000-image subset of the ImageNet~\cite{imagenet} training set. Top-1 accuracies for perturbation-free images are 0.58, 0.69 for CaffeNet and VGG-16, respectively. Similarly, top-1 accuracies for adversarially perturbed images with no noise masking are 0.1 and 0.25 for CaffeNet and VGG-16, respectively. For VGG-16, masking the noise in just 50\% of the ranked filter activations restores more than $\approx$ 80\% of the baseline accuracy on perturbation-free images.}   
	\label{fig:masking_other}
\end{figure*}

\section{Masking Perturbations in Other Layers}
In Section~\ref{subsec:filter_search} of the main paper (Figure~\ref{fig:masking} in the main paper), we evaluate the effect of masking $\ell_\infty$-norm adversarial perturbations in a ranked subset (using $\ell_1$-norm ranking) of convolutional filter activation maps of the first convolutional layer of a DNN. Here, in Figure~\ref{fig:masking_other}, we evaluate the effect of masking $\ell_\infty$-norm adversarial perturbations in ranked filter activation maps of the convolutional layers 2, 3, 4 and 5 of CaffeNet~\cite{AlexNet} and VGG-16~\cite{vggnet}. We use the same evaluation setup as in Section~\ref{subsec:filter_search} of the main paper (i.e., 1000 image random subset of the ImageNet~\cite{imagenet} training set). The top-1 accuracy for perturbation-free images of the subset are 0.58 and 0.69 for CaffeNet and VGG-16, respectively. Similarly, the top-1 accuracies for adversarially perturbed images in the subset are  0.10 and 0.25 for CaffeNet and VGG-16, respectively. Similar to our observations in Section~\ref{subsec:filter_search} of the main paper, for most DNN layers, masking the adversarial perturbations in just the top 50\% most susceptible filter activation maps (identified by using the $\ell_1$-norm ranking measure, Section~\ref{subsec:filter_search} of the paper), is able to recover most of the accuracy lost by the baseline DNN (Figure~\ref{fig:masking_other}). Specifically, masking the adversarial perturbations in the top 50\% ranked filters of VGG-16 is able to restore at least 84\% of the baseline accuracy on perturbation-free images.

\section{\emph{Feature Regeneration Units}: An Ablation Study}
\begin{figure}[!t]
	\centering
    \includegraphics[width=0.47\textwidth]{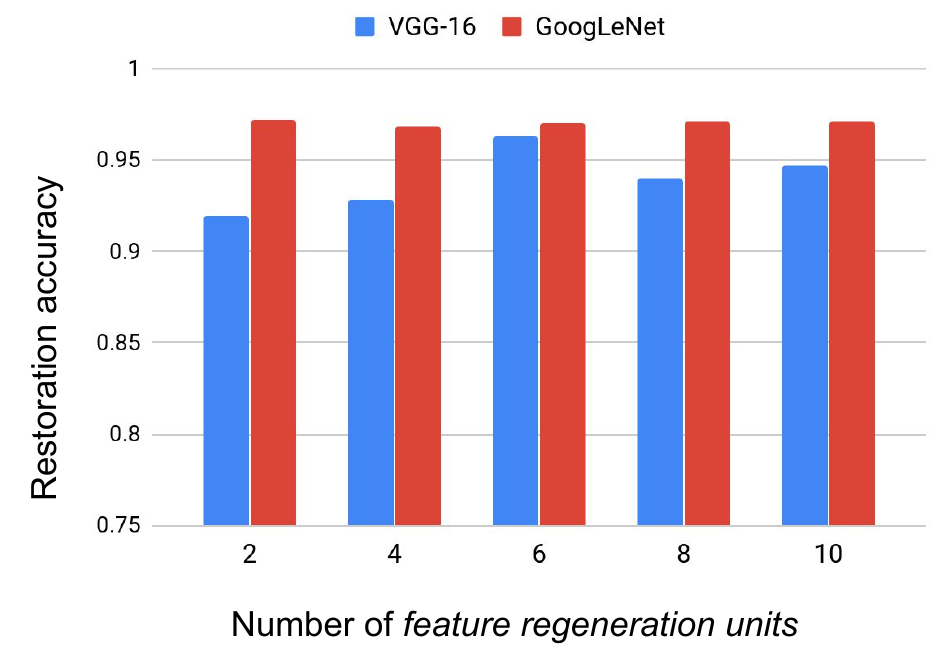}
	\caption{Effect of adding \emph{feature regeneration units} on the restoration accuracy of our proposed defense. Adding just two \emph{feature regeneration units} in GoogLeNet~\cite{GoogLeNet} achieves a restoration accuracy of 97\% and adding more \emph{feature regeneration units} to the DNN does not improve results any further. For VGG-16~\cite{vggnet}, adding 6 \emph{feature regeneration units} provides best results.}   
	\label{fig:ablation}
\end{figure}
In general, \emph{feature regeneration units} can be added at the output of each convolutional layer in a DNN. However, this may come at the cost of increased computations, due to an increase in the number of DNN parameters. As mentioned in Section~\ref{subsec:defense_train} of the main paper, we constrain the number of \emph{feature regeneration units} added to the DNN, in order to avoid drastically increasing the training and inference cost for larger DNNs (i.e., VGG-16, GoogLeNet and ResNet-152). Here, we perform an ablation study to identify the least number of \emph{feature regeneration units} needed to at least achieve a 95\% restoration accuracy across most DNNs. Specifically, we use VGG-16~\cite{vggnet} and GoogLeNet~\cite{GoogLeNet} for this analysis. We evaluate the restoration accuracy on the ImageNet~\cite{imagenet} validation set (ILSVRC2012) by adding an increasing number of \emph{feature regeneration units}, starting from a minimum value of 2 towards a maximum value of 10 in steps of 2. Starting from the first convolutional layer in a DNN, each additional \emph{feature regeneration unit} is added at the output of every second convolutional layer. In Figure~\ref{fig:ablation}, we report the results of this ablation study and observe that for GoogLeNet, adding just two \emph{feature regeneration units} achieves a restoration accuracy of 97\% and adding any more \emph{feature regeneration units} does not have any significant impact on the restoration accuracy. However, for VGG-16, adding only 2  ~\emph{feature regeneration units} achieves a restoration accuracy of only 91\%. For VGG-16, adding more \emph{feature regeneration units} improves the performance with the best restoration accuracy of 96.2\% achieved with 6 \emph{feature regeneration units}. Adding more than 6 \emph{feature regeneration units} resulted in a minor drop in restoration accuracy and this may be due to data over-fitting. As a result, we restrict the number of \emph{feature regeneration units} deployed for any DNN to $\min(\#\text{DNN layers}, 6)$.

\begin{table}[!t]
\caption{Defense restoration accuracy for oracle DNNs equipped with our defense for an $\ell_\infty$-norm UAP~\cite{UAP} attack ($\xi = 10$) using surrogate defense DNNs equipped with our defense.}
\vspace{4pt}
\centering
\resizebox{0.49\textwidth}{!}{
\renewcommand{\arraystretch}{1.1}%

\begin{tabular}{l|ccc}
\hline
\multicolumn{1}{c|}{\multirow{2}{*}{\textbf{Surrogate}}} & \multicolumn{3}{c}{\textbf{Oracle}} \\ \cline{2-4} 
\multicolumn{1}{c|}{} & \multicolumn{1}{l}{\textbf{VGG-F + defense}} & \multicolumn{1}{l}{\textbf{GoogLeNet + defense}} & \multicolumn{1}{l}{\textbf{VGG-16 + defense}} \\ \hline
\textbf{CaffeNet + defense} & 0.906 & 0.963 & 0.942 \\
\textbf{Res152 + defense} & 0.889 & 0.925 & 0.925 \\ \hline
\end{tabular}
}
\vspace*{-2ex}
\label{table:transfer}

\end{table}

\begin{figure}[!t]
	\centering
    \subfloat{\hspace{-1ex}\includegraphics[width=0.49\textwidth]{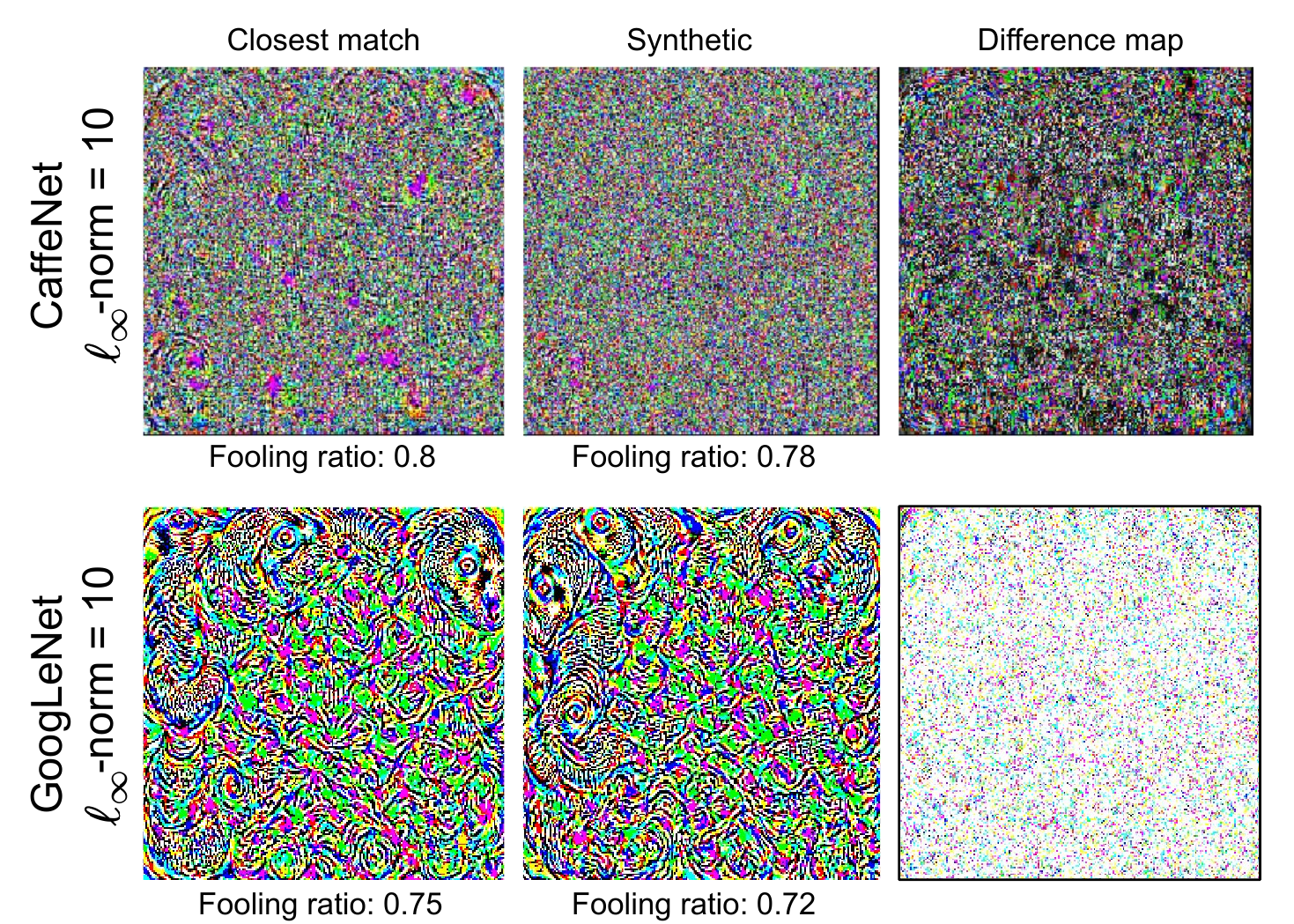}}
	\caption{Visualization of synthetic perturbations (center) computed for CaffeNet~\cite{AlexNet} and GoogLeNet~\cite{GoogLeNet} along with their closest match in the set of original perturbations (left) and a per pixel difference map between the two (right).}   
	\label{fig:_syn}
\end{figure}

\section{Attacks using Surrogate Defense DNNs}
\label{subsubsec:surrogate}

In this section, we evaluate if it is possible for an attacker/adversary to construct a surrogate defense network if it was known that our defense was adopted. In situations where exact defense (\emph{feature regeneration units} + baseline DNN) is typically hidden from the attacker (\emph{oracle}), a DNN predicting output labels similar to our defense (\emph{surrogate}), can be effective only if an attack generated using the \emph{surrogate} is transferable to the \emph{oracle}. UAP~\cite{UAP} attacks are transferable across baseline DNNs (Table~\ref{table:cross-dnn} in main paper), i.e., adversarial perturbation computed for a DNN whose model weights and architecture are known (surrogate) can also effectively fool another target DNN that has a similar prediction accuracy, but whose model weights and architecture are not known to the attacker (oracle). Assuming that our defense (\emph{feature regeneration units} + baseline DNN) for CaffeNet~\cite{AlexNet} and Res152~\cite{he2016deep} is available publicly as a \emph{surrogate}, universal attack examples computed from these DNNs may be used to attack our defenses for other DNNs, e.g. VGG-F or VGG-16 as an \emph{oracle}. We show in Table~\ref{table:transfer} that our defense mechanism successfully breaks attack transferability and is not susceptible to attacks from \emph{surrogate} DNNs based on our defense.

\section{Examples of Synthetic Perturbations}
Sample visualizations of synthetic adversarial perturbations generated using our algorithm proposed in Section~\ref{subsec:synthetic_data}~(Algorithm~\ref{alg:gen_sample}) of the main
paper are provided in Figure~\ref{fig:_syn}.

\section{Examples of Feature Regeneration}
\begin{figure*}[]
	\centering
	\vspace*{-2ex}
    \subfloat{\hspace{-2ex}\includegraphics[width=.98\textwidth]{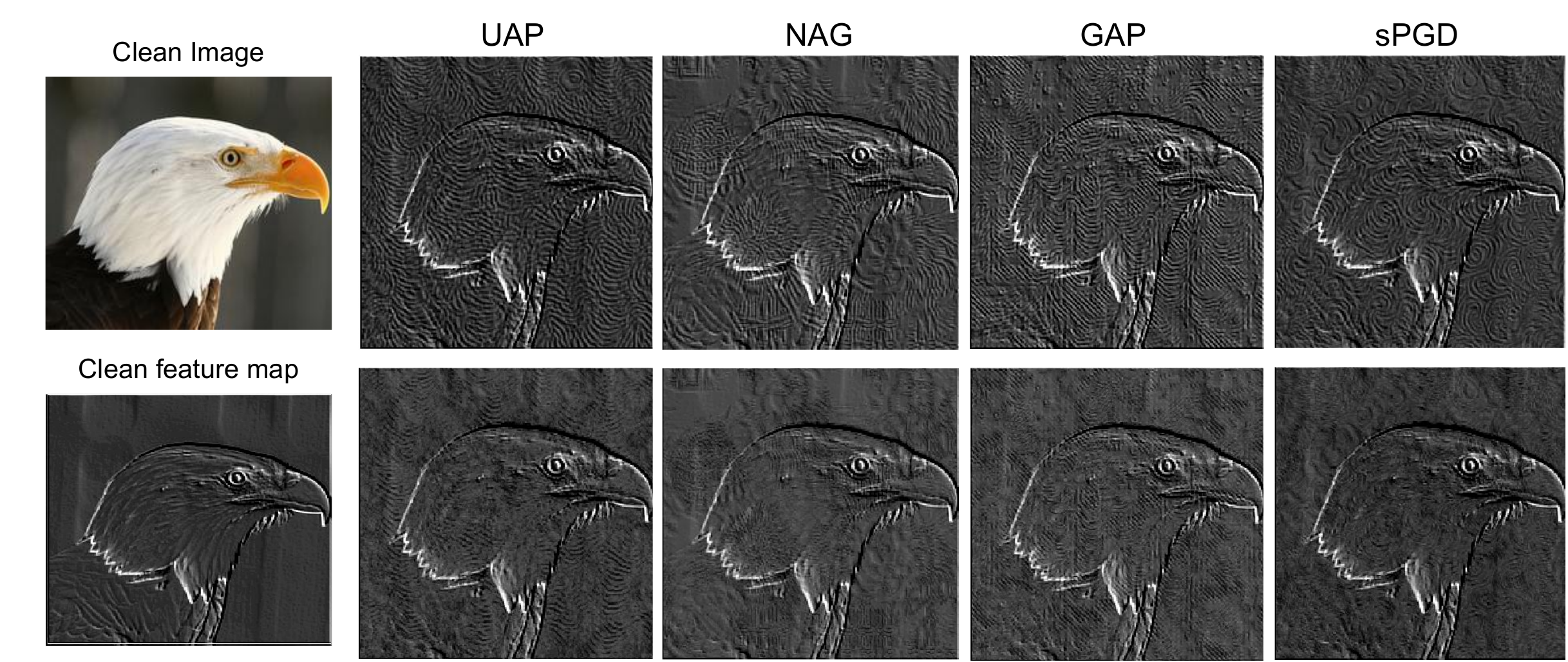}}
    \vspace*{-2ex}
    \subfloat{\hspace{-2ex}\includegraphics[width=0.98\textwidth]{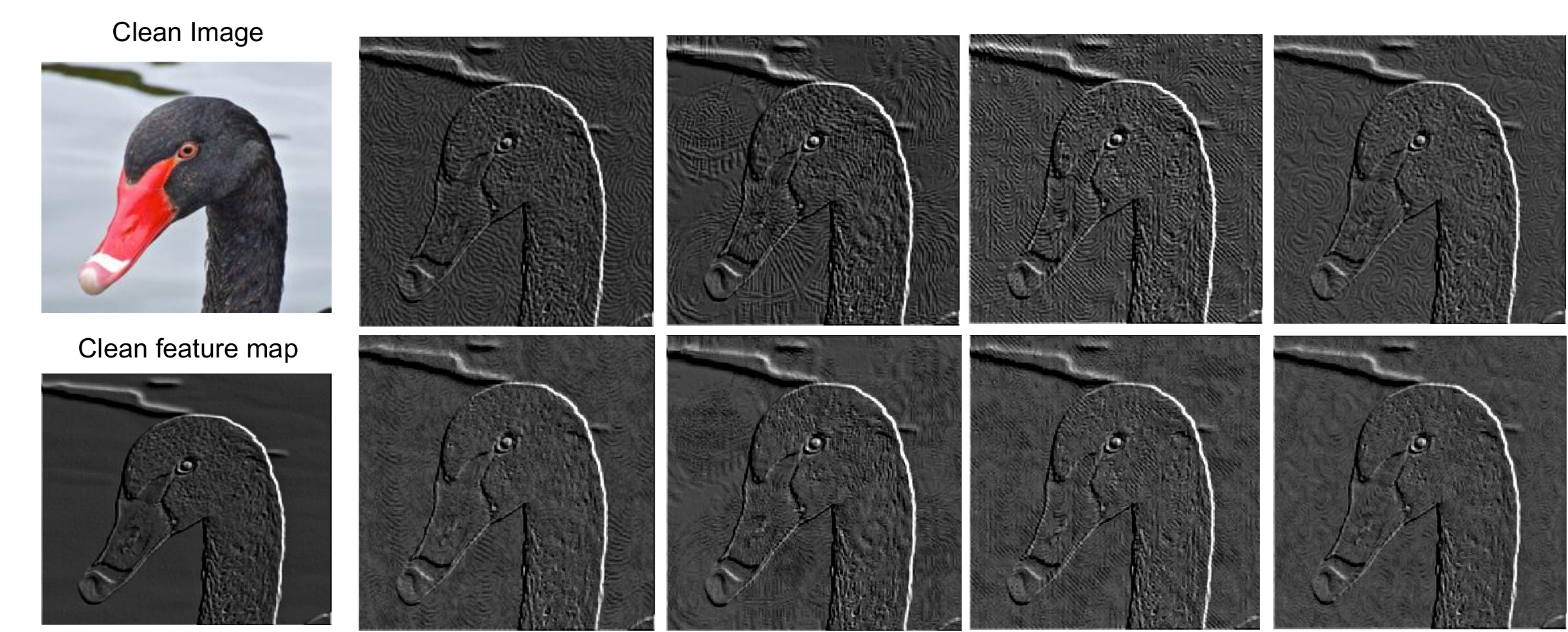}} 
    \vspace*{-2ex}
    \subfloat{\hspace{-2ex}\includegraphics[width=0.98\textwidth]{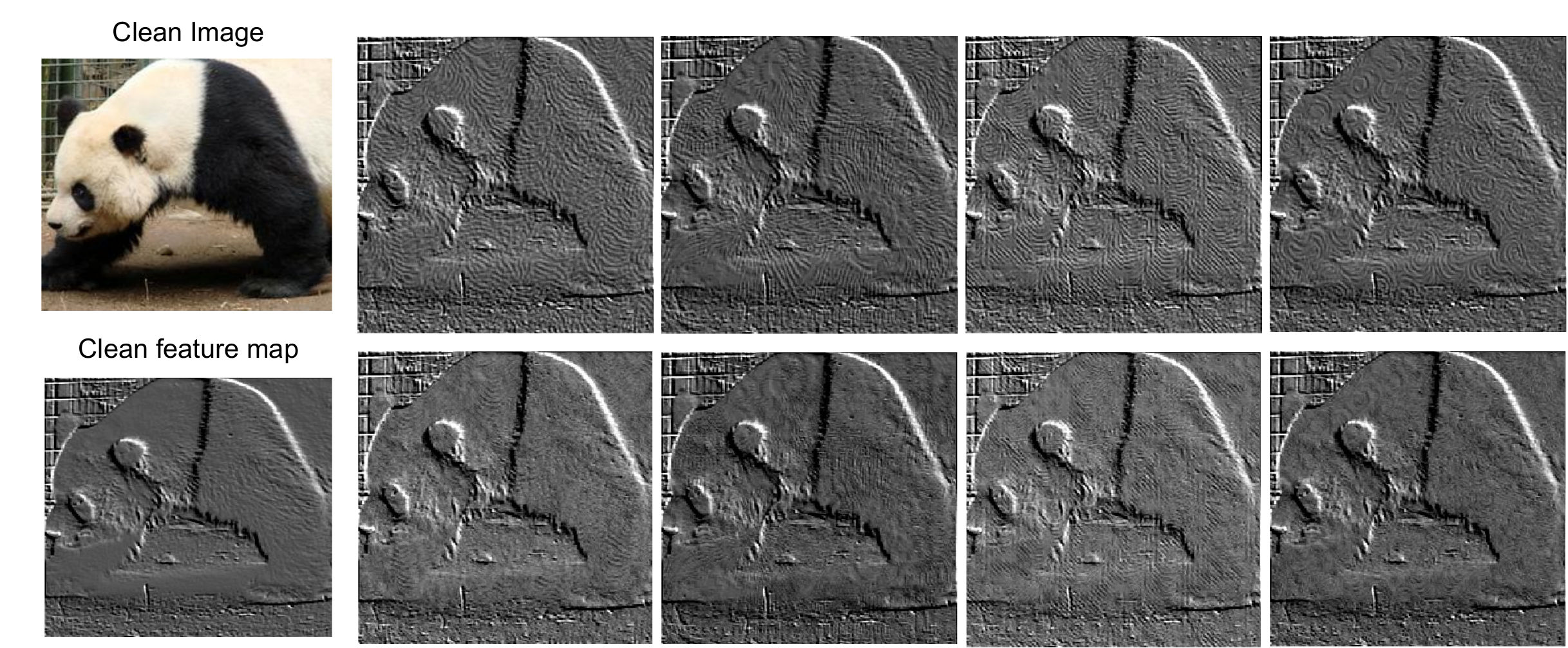}}
	\caption{Visual examples of DNN feature maps before and after \emph{feature regeneration} using our proposed method, for images perturbed by universal perturbations~(UAP~\cite{UAP}, NAG~\cite{nag}, GAP~\cite{gap} and sPGD~\cite{spgd}). Perturbation-free feature map (clean feature map), different adversarially perturbed feature maps (Rows 1, 3 and 5) and corresponding feature maps regenerated by \emph{feature regeneration units} (Rows 2, 4 and 6) are obtained for a single filter channel in conv1\_1 layer of VGG-16~\cite{vggnet}. Our \emph{Feature regeneration units} are only trained on UAP~\cite{UAP} attack examples.   }
	\label{fig:feat_corr_add}
\end{figure*}

Additional visualizations of DNN feature maps before and after \emph{feature regeneration} using our proposed defense in Section~\ref{subsec:defender_unit} of the main paper are provided in Figure~\ref{fig:feat_corr_add}.

\section{Examples of Universal Attack Perturbations}
\begin{figure*}[]
	\centering
    \subfloat{\includegraphics[width=0.93\textwidth]{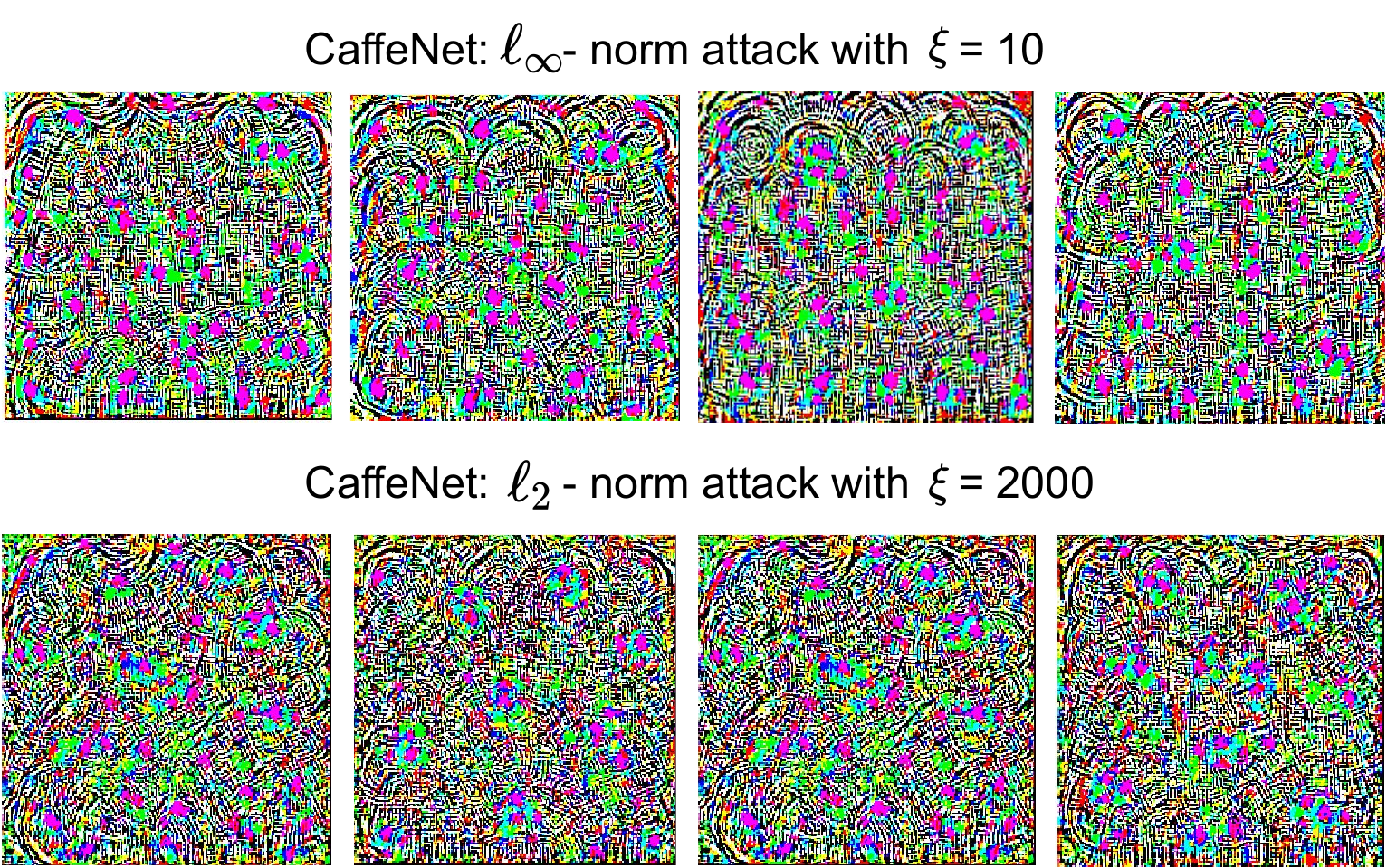}}
    \hfill
    \subfloat{\includegraphics[width=0.93\textwidth]{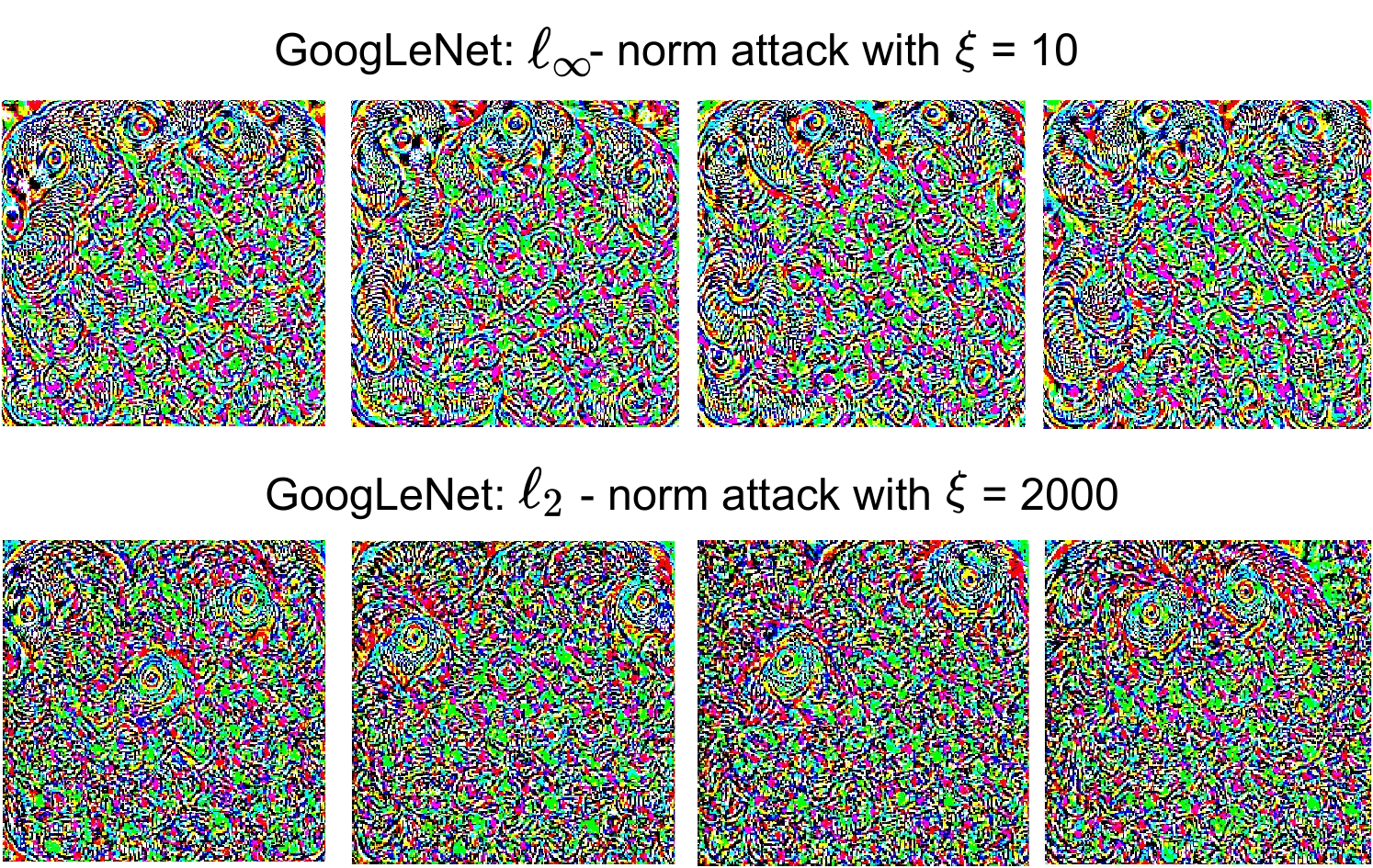}} 
	\caption{Visual examples of $\ell_\infty$-norm and $\ell_2$-norm UAP~\cite{UAP} attack test perturbations for CaffeNet~\cite{AlexNet} and GoogLeNet~\cite{GoogLeNet}.}  
	\label{fig:_uap}
\end{figure*}

Sample visualizations of $\ell_\infty$-norm and $\ell_2$-norm UAP~\cite{UAP} attack perturbations are shown in Figure~\ref{fig:_uap}.

\end{document}